# Can predictive models be used for causal inference?

Maximilian Pichler[1,*], Florian Hartig[1]

[1] Theoretical Ecology, University of Regensburg, Universitätsstraße 31, 93053 Regensburg, Germany

* corresponding author, maximilian.pichler@biologie.uni-regensburg.de

## Abstract

Supervised machine learning (ML) and deep learning (DL) algorithms excel at predictive tasks, but it is commonly assumed that they often do so by exploiting non-causal correlations, which may limit both interpretability and generalizability. Here, we show that this trade-off between explanation and prediction is not as deep and fundamental as expected. Whereas ML and DL algorithms will indeed tend to use non-causal features for prediction when fed indiscriminately with all data, it is possible to constrain the learning process of any ML and DL algorithm by selecting features according to Pearl's backdoor adjustment criterion. In such a situation, some algorithms, in particular deep neural networks, can provide near unbiased effect estimates under feature collinearity. Remaining biases are explained by the specific algorithmic structures as well as hyperparameter choice. Consequently, optimal hyperparameter settings are different when tuned for prediction or inference, confirming the general expectation of a trade-off between prediction and explanation. However, the effect of this trade-off is small compared to the effect of a causally constrained feature selection. Thus, once the causal relationship between the features is accounted for, the difference between prediction and explanation may be much smaller than commonly assumed. We also show that such causally constrained models generalize better to new data with altered collinearity structures, suggesting generalization failure may often be due to a lack of causal learning. Our results not only provide a perspective for using ML for inference of (causal) effects but also help to improve the generalizability of fitted ML and DL models to new data.

# Introduction

In the fields of statistics and machine learning, it is widely recognized that there is a difference between predictive and explanatory or causal modelling (1). One of the reasons is that using correlations between features and the response can improve predictions, even when those variables are not causally connected. Along with the bias-variance tradeoff (e.g. ((2, 3), the ability to exploit non-causal correlations likely explains the success of supervised machine learning (ML) and deep learning (DL) algorithms in predictive tasks (4–6); however, such a predictive modelling strategy tacitly accepts that trained ML models will in general not learn the true underlying relationships, which limits their interpretability (suggesting a prediction-explanation trade-off) and may also partly explain why they often do not generalize well to new data (suggesting an interpolation-extrapolation trade-off).

Specialized ML approaches for estimating causal effects exist (e.g. causal forest (7), double/debiased ML (8), metalearners (9) or causal discovery algorithms (10, 11)), and we will discuss the relationship between these and the present study later. Here, our goal is to understand if and when classical ML algorithms can correctly adjust for collinear features, which is a prerequisite for using them for in causal inference. By means of that, we can also explore if there is indeed a fundamental trade-off between prediction and explanation when training ML and DL algorithms (1).

The key idea of our study is that if the causal graph is known, research in causal inference has solved the problem of how we should select features such that a statistical regression model (e.g. ordinary least squared (OLS)) would adjust for confounding such that the causal effect of one or several target variables is correctly estimated (12). As pointed out by (13), these ideas should in principle be transferable to ML and DL models. However, given that ML and DL models rely heavily on (adaptive) regularization and induced regularization biases can affect causal estimates (14), it remains an open question how well this idea works in practice. Here, we address this problem by first suggesting an explainable AI (xAI) metric to extract effect estimates from fitted ML and DL models, and then performing a number of simulations to examine bias on effect estimates under collinearity in different ML and DL models.

## Causally constrained ML requires unbiased learning

To understand why bias in effect estimates is crucial for causal inference, we shortly summarize the general approach to separate correlation from causality in static data. The key problem is that a correlation may be caused by a direct causal link, but also by a third variable that causally influences both the feature of interest and the response (i.e., a confounder, see Table. 1). To adjust for the effect of such additional variables, one must first generate a hypothesis about the underlying graph which describes causal relationships between all features (12, 15). Based on this graph, one can isolate the underlying causal effect of the target feature by conditioning on the other features (adjustment), for example using multiple regressions or (piecewise) structural equation models (Table. 1, (16) see also (12)). As pointed out by (13), we should be able to transfer the same idea to ML and DL models. We refer to ML models trained with such a set of causally selected features as "causally constrained".

**Table 1: Common causal structures and their statistical adjustments:** The column DAG (short for: directed acyclical graph) describes the assumed causal relationship between the variables. The columns "description" describes the correlations created by the respective relationships and the usual statistical adjustment. The estimated effects (raw: P(C|A) and adjusted: P(C|A,B) in a multiple regression) are visualized in the last column to the right.

| DAG | Description | P(C\|A) | P(C\|A, B) |
|---|---|---|---|
| Confounder<br>B → A, B → C<br>A ⇢ C | • If we look at the unconditional correlation P(C\|A), we see a spurious effect.<br>• By conditioning on the confounder P(C\|A,B), we can isolate the true causal effect | flat line at C=0 over A∈[-1,+1] | positive linear line from (-1,-1) to (+1,+1) over A\|B |
| Mediator<br>A → B → C<br>A ⇢ C | • If we look at the unconditional correlation P(C\|A), we see the total effect<br>• By conditioning on the mediator P(C\|A,B), we can isolate the direct causal effect | flat line at C=0 over A∈[-1,+1] | positive linear line from (-1,-1) to (+1,+1) over A\|B |
| Collider<br>A → B, C → B<br>A ⇢ C | • If we look at the unconditional correlation P(C\|A), we see the true causal effect.<br>• By conditioning on a collider p(C\|A,B), we create a collider bias and obtain the wrong causal effect | positive linear line from (-1,-1) to (+1,+1) over A | flat line at C=0 over A\|B∈[-1,+1] |

This argument assumes, however that ML algorithms (similar to ordinary least squared (OLS) regression) provide unbiased effect estimates under collinearity so that the adjustment sketched in Table 1 can remove the entire effect of possible confounders, and there are several reasons to cast doubt on that assumption.

Most importantly, it is well-known that certain ML techniques trade off bias against variance, which can disproportionally bias collinear feature effects. For example, shrinkage estimators such as LASSO, RIDGE or elastic-net, although originally motivated by the desire to improve

OLS estimates under collinearity (18), tend to push strong effects of a feature over to other collinear features where the shrinkage loss is weaker (Fig. 1, S3) (14). This creates a stronger regularization bias for collinear features than for independent features or the predictions. We call this phenomenon that a causal effect moves over to collinear non-causal feature a "causal spillover".

Similar issues may arise in ensemble models. In the popular random forest (RF) algorithm, for example, variance in the tree ensemble is increased by randomly hiding features at each split of each tree (19). This variance decreases the correlation between ensemble members, which can reduce the predictive error of the ensemble (20–22). However, if a confounder is hidden by this process, its casual effect will spill over to other collinear features, inducing a bias in the effect estimates (23, confirmed by Fig. 1).

A different effect can occur in greedy learning algorithms such as (gradient) boosted regression trees (BRT). In these algorithms, weaker collinear features are only used when the stronger ones are exhausted, which can occur within the internal regression trees or could potentially arise from the boosting (Fig. 1, Fig. S1). Based on this, there is a concern that strong features steal effects from weaker collinear features (we refer to this as "causal greediness").

For neural networks (NN), it is unclear if such biases are expected. Pure (deep) NNs do not explicitly include any of the previously mentioned regularization mechanisms. Nevertheless, it is often reported that trained NNs display a simplicity bias akin to an implicit regularization, which has been associated to the stochastic gradient descent or network architecture (24, 25). Such a simplicity bias could lead to similar causal spillovers as those reported for the elastic net. More importantly, however, NN are in practice usually trained with additional regularization, for example in the form of shrinkage (e.g., elastic net) or dropout. The latter implicitly creates an ensemble model (26) and could lead to similar causal spillover as in random forest (Fig. S3).

## Measuring causal bias of trained models via xAI

A complication for quantifying and comparing to what extent these theoretical considerations apply for trained ML and DL models is that those models do not directly report effect sizes. However, it is possible to extract feature effects using appropriate model-agnostic explainable AI (xAI) method (27). A large number of xAI methods exist, but most quantify feature importance for predictions, which is more analogous to variance partitioning in an ANOVA setting (i.e. a joint measure of effect and variance of a feature) and not of effect sizes (28).

Moreover, it is know that many xAI metrics are not robust against feature collinearity, for example because univariate unconditional permutations will generate feature combinations that are outside the range of collinear data (29–32), which makes them unreliable for our purpose.

In search for a model-agnostic post-hoc xAI metric that corresponds, in the case of linear effects of possible collinear features, exactly to effect sizes estimated by linear regression, we settled on the idea of average conditional effects (ACEs). ACEs, in the statistical literature also known as average marginal effects, are a common choice to extract average effects for nonlinear statistical models (33). The basic idea behind the ACE is to use the fitted model and calculate the average local derivative of the prediction with respect to a target feature over all observations (see methods). With *n* observations, the $ACE$ for vector $x_k$ (*k* indexing the features) is then:

$$ACE_{x_k} = \frac{1}{n} \sum_{i=1}^{n} \frac{\partial \hat{f}(X)}{\partial x_k^{(i)}}$$

The idea to use ACE to interpret ML model is not new (34), but whereas (34) suggested to extend ACE for non-linear effects by splitting the feature space in different regions, we argue that a robust average across feature-output relationship corresponds exactly to what one would visually characterize as a learned causal effect and what we need here to compare our ML algorithms with the OLS. The ACE can also easily be extended to infer two-way or higher interactions (see SI Appendix 1.2).

We acknowledge that there are alternatives to ACEs, in particular global xAI metrics based on Shapley values (35), their algorithmically specific versions such as kernSHAP or treeSHAP values (36, 37), or accumulated local effect plots (38). But those do not map directly on regression slopes and are computationally expensive, whereas our results show that ACEs are fast to compute and correspond very well to OLS effect sizes in a linear simulation setting.

## Results

Equipped with an xAI method for extracting main effects and interactions from fitted ML models, we proceed to examine if ML models learn unbiased effects under feature collinearity. We considered the four major classes of ML algorithms currently in use, which are also representative of different ML paradigms: Random forest (RF, using the bagging paradigm) and boosted regression trees (BRT, using the boosting paradigm) are both ensemble models that rely on the principles of model averaging (21, 22), deep neural networks are representatives of artificial neural networks (NN, 3 hidden layers with 50 hidden nodes), and elastic-net regression models with a LASSO and Ridge regularization (paradigm of shrinkage estimators) (14).

## Near-asymptotic performance

In a large-data situation (see methods), our results confirm the theoretical expectations that RF as well as to a lesser degree BRT and elastic net are principally biased towards smaller effect sizes (regularization), even if there is no collinearity (Fig. 1, row a), whereas pure NN is near unbiased. We also included an OLS regression as a reference, which is mathematically known to be unbiased.

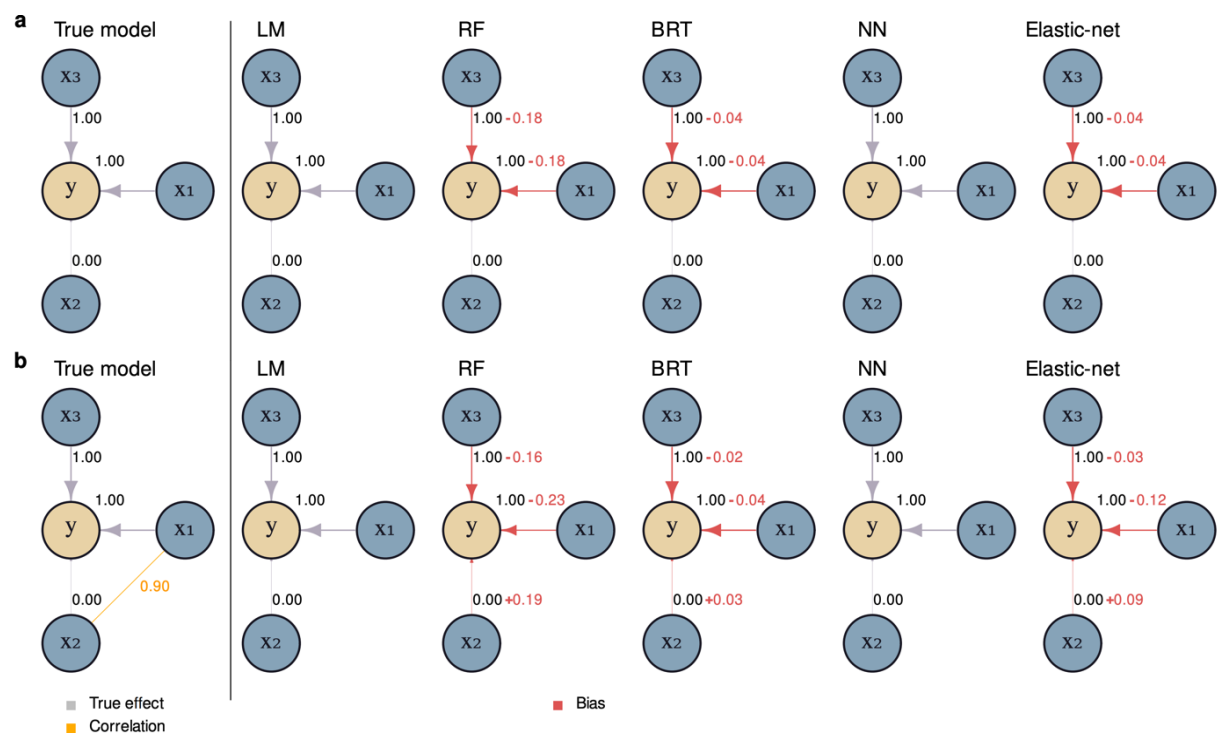

**Figure 1:** Quantification of causal biases and spillover for different ML algorithms when trained on data simulated from two different causal relationships (a: uncorrelated features with effect sizes *($\beta_1 = 1.0$,  $\beta_2 = 0.0$, and $\beta_3 = 1.0$)*, b: $x_1$ and $x_2$ being strongly correlated (Pearson correlation factor = 0.9) but only $x_1$ affects *y.*). Sample sizes were sufficiently large that stochastic effects can be excluded (1000 observations and 500 repetitions). Effects of the ML models were quantified using average conditional effects.

Under collinearity (Fig. 1, rows b), additional algorithm-specific biases arise: the strongest causal spillover is observed for the RF algorithm, presumably because feature subsampling leads to open backdoors (already shown by 23), followed by elastic-net and BRT, whereas the NN remained unbiased. The fact that BRT showed light spillover and no causal greediness was against our expectation. We explored further and found that pure linear boosting only initially leads to causal greediness; however, in the course of further boosting steps, this is compensated, resulting in an unbiased effect estimate (Fig. S4). The spillover that we observe is likely caused by other features of the boosting algorithm, in particular the use of regression trees (Fig. S3).

We note that the NN was trained without any regularization, which may be considered unrealistic in a practical scenario. An NN trained with regularization via dropout (39) (rate = 0.3) showed similar biases as RF and elastic-net (Fig. S5). We explain this by the fact that dropout, similar to the feature subsampling in RF, hides some effects during training which can lead to causal spillover.

## Performance in data-poor situations

While our previous results using a large sample size allow us to understand the mechanisms by which ML algorithms introduce bias into effect estimates, they may also seem somewhat discouraging because, except for the NN, all ML models perform considerably worse than a simple linear regression (Fig. 1). This, however, was to be expected, because OLS is known to be the best linear unbiased estimator (BLUE) for estimating feature effects.

Advantages for ML models over OLS are expected when either the functional form of the response is nonlinear or unknown, or when there is an advantage to be gained from trading off bias against variance, which is the case in data-poor situations when the variance contributes significantly to the total error ($MSE = Bias^2 + Var + \sigma^2$ with $\sigma^2 = irreprodcible\ error$). In such a situation, ML algorithms might outperform OLS in estimating complex or nonlinear effects, in particular if model hyperparameters that adjust the regularization strength are tuned. To examine such a scenario, we simulated a data-poor regression situation with 100 features and 50, 100, and 600 observations (see methods).

### The impact of hyperparameters and a separate bias-variance tradeoff for inference and predictions

In such a data-poor situation, we expect that hyperparameters need to be tuned, and we expect that there is a tradeoff between tuning for either explaining or predicting. To test this, we sampled 1000 different hyperparameters for each model (Table S1), calculated the bias and variance for effects and predictions (20 replicates), and modeled the effects of hyperparameters on predictive and inferential MSE using generalized additive models (GAM) and random forest (RF).

We find that hyperparameters have significant effects on both bias and variance of effect estimates and predictions. For NNs, the SELU activation function caused the smallest bias on the effect estimate and the prediction (Fig. 2), but this effect decreased with increasing observations (Fig. S7-S9). More hidden layers (depth) increased the bias on the effect estimates and the prediction (Fig. 2). For BRT, larger learning rates (eta) and larger number of trees decreased bias on the effect estimates and predictors (Fig. 2). For RF, more features that are used in each split (mtry) and larger minimum node sizes decreased the biases (Fig. 2). For elastic net, as expected, alpha and lambda had strong effects on the effect and prediction errors (Fig. 2).

Often, the effects of the hyperparameters on bias and variance were contrary (depth in NN, eta in BRT, and mtry in RF), which reflects the well-known bias-variance tradeoff and explains why the optimal set of hyperparameters (red, predicted by a RF, Fig. 2) is not at the marginal optima of the hyperparameter-error associations (Fig. 2).

Most importantly, although the bias-variance tradeoffs for inference and prediction often showed similar tendencies for hyperparameters, some hyperparameters had notably different effects for the two goals (Fig. 2), for example the number of features to select from in RF (mtry in RF). That and the fact that the variance was on different scales for effect estimate and prediction error (not shown in Fig. 2) led to different optimal hyperparameter sets, meaning that even if the models are causally constrained via the feature selection, there is a trade-off between tuning their hyperparameters for predictions or for inference.

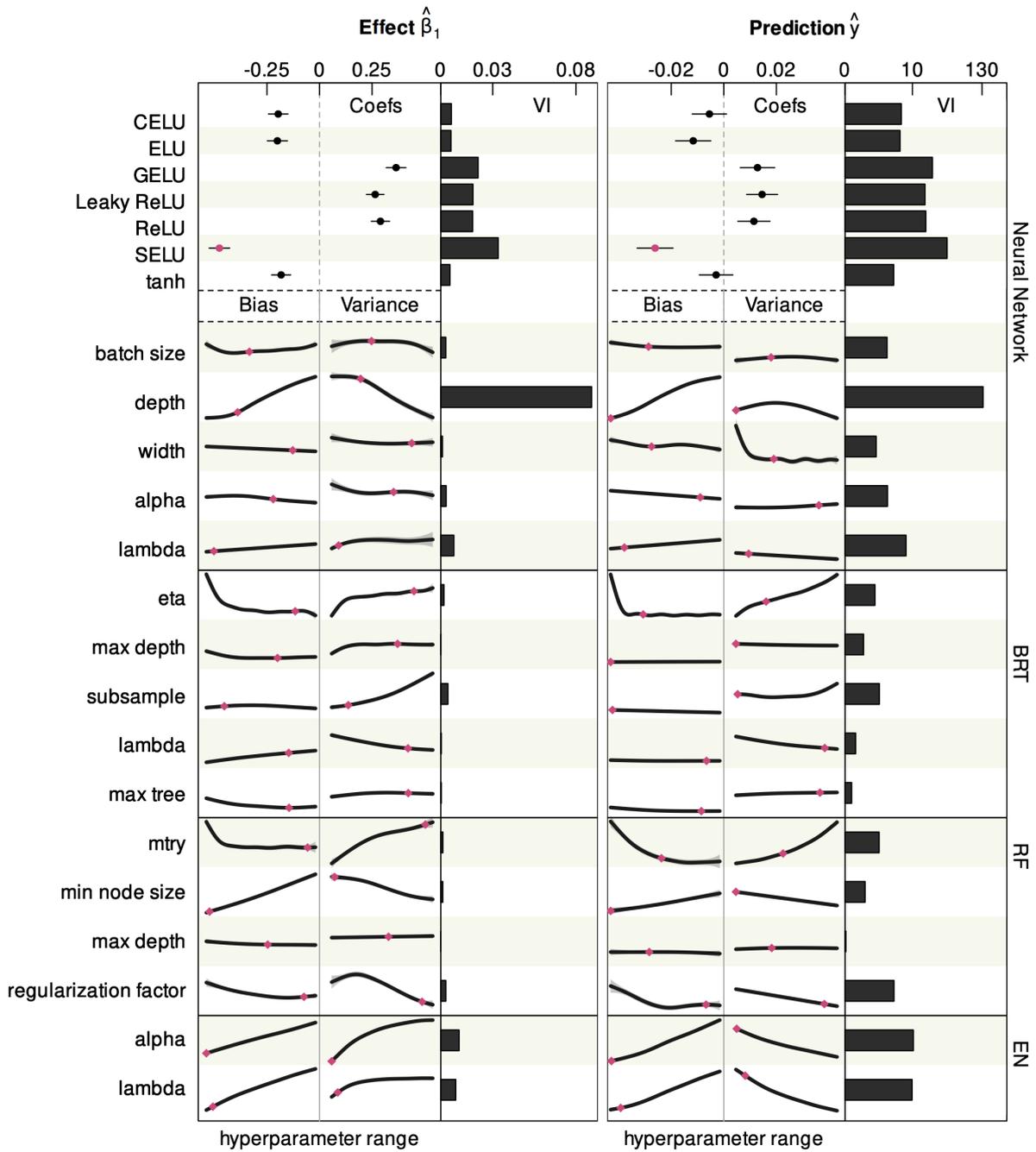

Figure 2: Results of hyperparameter tuning for Neural Networks (NN), Boosted Regression Trees (BRT), Random Forests (RF), and Elastic Net (EN) for 100 observations with 100 features. The influence of the hyperparameters on effect $\hat{\beta}_1$ (bias, variance, and MSE), and the predictions of the model, $\hat{y}$, (bias, variance, and MSE) were estimated by a multivariate generalized additive model (GAM). Categorical hyperparameters (activation function in NN) were estimated as fixed effects. The responses (bias, variance, MSE) were centered so that the categorical hyperparameters correspond to the intercepts. The variable importance of the hyperparameters was estimated by a random forest with the MSE of the effect $\hat{\beta}_1$ (first plot) or the prediction (second plot) as the response. Red dots correspond to the best predicted set of hyperparameters (based on a random forest), in the first plot for the minimum MSE of the effect for $\hat{\beta}_1$ and in the second plot for the minimum MSE of the predictions ($\hat{y}$).

# Bias and error on effects induced by algorithm and hyperparameter choice in data-poor simulations

Based on the optimal hyperparameters for prediction and inference, we then quantified bias and variance of present or absent feature effects under feature collinearity (see methods) for all algorithms and again OLS as a reference. Our results show that, as expected, all ML algorithms apply regularization, resulting in increasing bias with smaller data sizes (Fig. 2). Relatively, however, NN and elastic-net showed the smallest biases, which decreased stronger with more observations while RF showed the largest biases. For the second effect estimate, the zero effect ($\beta_2$), all models showed small biases (Fig. 3). For 600 observations, the LM was unbiased, as expected (Fig. 3). Variance was small for all effect estimates (Fig. 3). Given that elastic net is not a general function approximator but rather "just" a regularized OLS model, we conclude that of the general algorithms, NN seems the preferable choice for the purpose of causal inference.

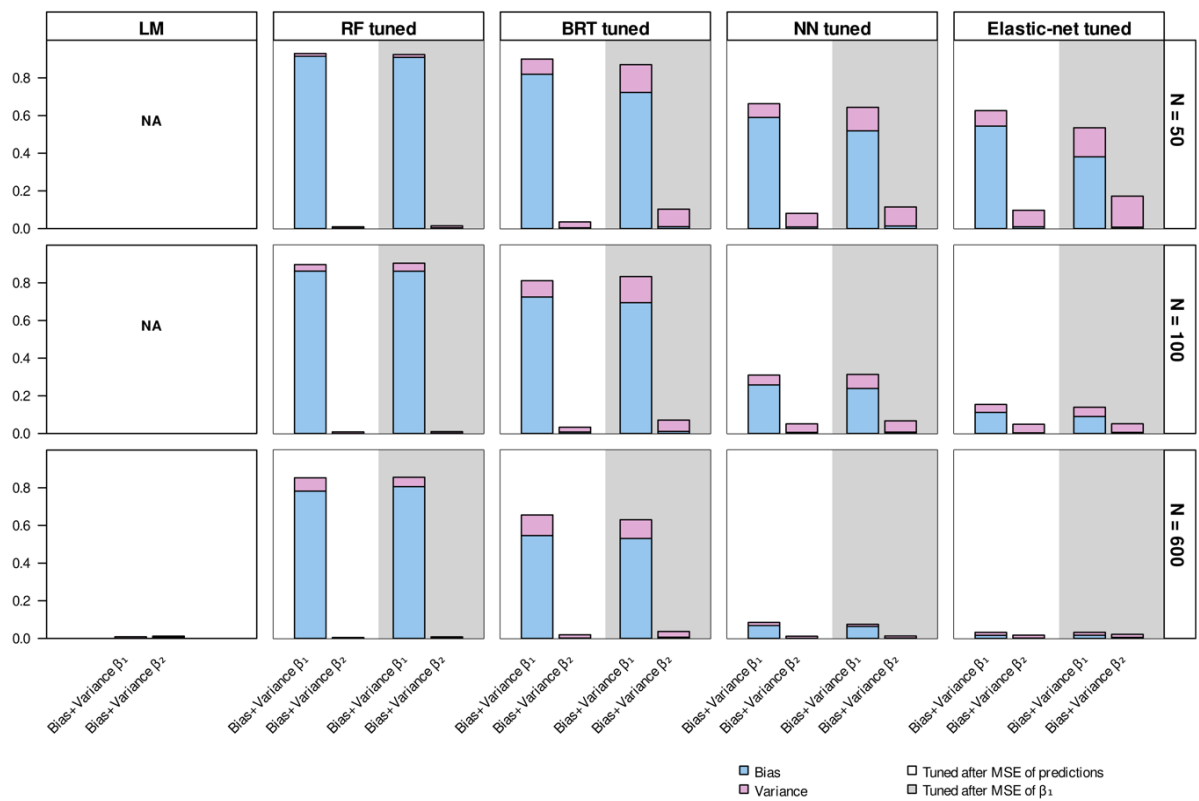

**Figure 3:** Bias and variance of estimated effects in data-poor situations. N = 50, 100, and 600 observations of 100 weakly correlated features were simulated. True effects in the data generating model were $\beta_1 = 1.0$, $\beta_2 = 0.0$, and the other 98 effects were equally spaced between 0 and 1. Models were fitted to the simulated data (1000 replicates) with the optimal hyperparameters (except for LM, which doesn't have hyperparameters). Hyperparameters were selected based on the minimum MSE of $\hat{\beta}_1$ (green) or the prediction error (based on $\hat{y}$) (red). Bias and variance were calculated for $\hat{\beta}_1$ and $\hat{\bar{\beta}}_1$. Effects ($\hat{\beta}_i$ for $i = 1, \ldots, 100$) were approximated using ACE.

Models with hyperparameters tuned for inference had, on average, lower errors than when using hyperparameters tuned for prediction (Fig. 3, Table S2). This confirms that the bias-variance tradeoff is different for prediction and inference tasks.

## Case Study – Predicting out-of-distribution

Having seen that NN can infer near-unbiased estimates under collinearity (given enough observations), it seems contradictory that algorithms such as RF, with strongly biased effect estimates, often outperform NNs in predictive benchmarks. Such results, however, are usually obtained on test data that is out-of-sample but in-distribution, which means that the feature correlation in the hold-out data is identical to the training data. In such a case, we know a priori that the predictor $\hat{y}$ can be unbiased whilst having biased estimated effects $\hat{\beta}$ (causal spillover) (1, 40). If we predict out-of-distribution, however, for example to predict the effect of interventions or when the correlation structures change (e.g. latent confounders) (Fig. 4), it will be much less likely that a non-causal model delivers unbiased predictions (11, 41).

To demonstrate this phenomenon, we simulated a case study where we assumed that the goal is to predict lung cancer based on smoking and diet (Fig. 4) in two different scenarios. In this case study, we assume models are first trained and validated to predict lung cancer in an observational study and are then used to predict lung cancer in a randomized controlled trial (RCT) (Figure 5). By means of the control, the RCT forced predictions to be out-of-distribution. Specifically, we assume that the collider lung volume and the latent confounder financial constraints were controlled in the selection of trial participants, which means that they no longer correlate with the treatment. (Fig. 4). We trained three ML algorithms (RF, BRT, and NN) with two different feature selections, a conventional (full) model with all features (smoking, diet, and lung volume) and a causally constrained model with only smoking and diet used as features.

We find that in-distribution, the unconstrained model that uses all features outperformed the causally constrained models. In the second prediction scenario (out-of-distribution), however, the causally constrained models outperformed the conventional (full) model (Fig. 4). The reason is that including the collider in the training creates a collider bias which biases the effect of smoking. Without the collider, the causal effect of smoking is estimated with lower bias, which reduces the out-of-distribution prediction error (Fig. 4).

The case study also confirms our previous results that NN and BRT perform better than RF in estimating the true causal effects and thus in the out-of-distribution tasks. RF is unable to correctly separate collinear features (Fig. 1), leading to causal spillover in RF between diet and smoking during training (Fig. 4). However, RF achieves a lower prediction error for the full model (with collider) than BRT and NN, probably by chance because the causal spillover inadvertently leads to advantageous biases due to the collider (Fig. 4).

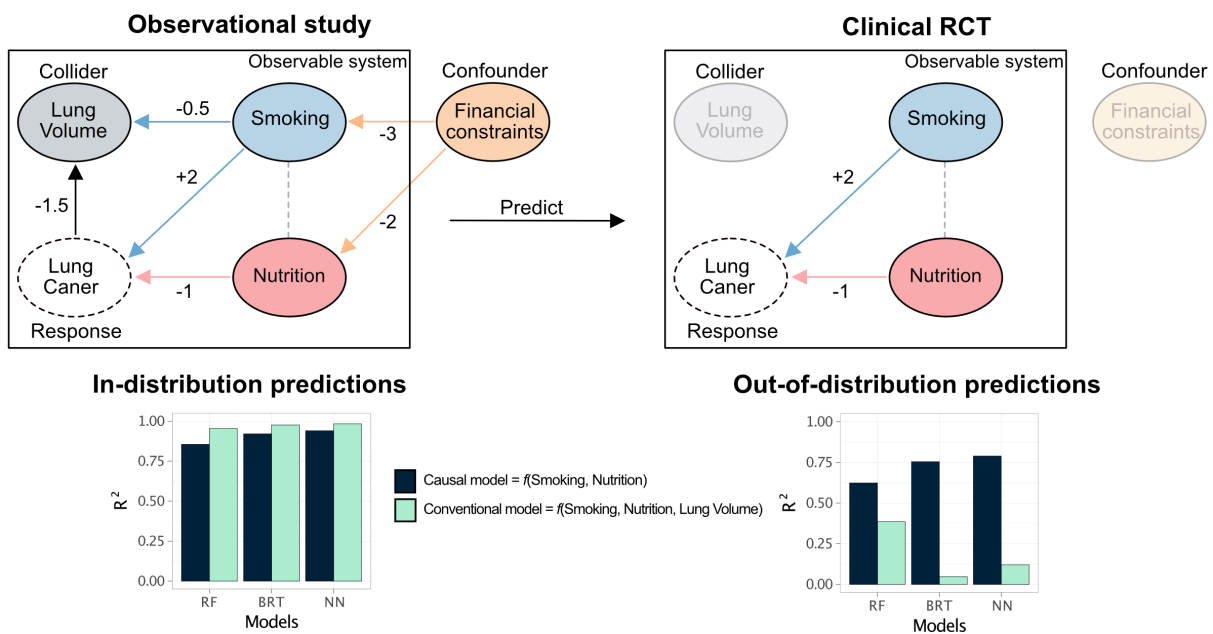

**Figure 4:** Difference between causal and conventional ML models for in-distribution and out-of-distribution predictions in a simulated case study. We assume that a first study collected data about the effects of smoking and nutrition on lung cancer. Lung volume is a collider and financial constraints is an unobservable confounder. Smoking and nutrition are correlated because of their latent confounder financial constraints. Then, the data from the observational study (left column) was used to train two different models, causally constrained (causal model) and a conventional model with all features, and to predict lung cancer in another observational study (in-distribution predictions) and in a clinical randomized controlled trial (RCT) (out-of-distribution predictions). In the RCT, patients were treated for lung volume, and received financial support (right side). Lung volume was removed as feature in the causal model because its inclusion would lead to biased effect estimates of smoking and nutrition (collider bias). Smoking and nutrition were both included to block the effect of the unobservable confounder on lung cancer (i.e., lung cancer and financial constraints are d-separated). In the first prediction scenario, the conventional model slightly outperformed the causal model (as measured by $R^2$), whereas in the second, out-of-distribution model, the causal model outperformed the conventional model.

# Discussion

The aim of our study was to understand if ML and DL algorithms display inherent biases, caused by algorithmic features and regularization methods, that prevents them to separate causal effects in the presence of feature collinearity. Our main finding is that this is indeed partly the case, but not to the same extent for all algorithms and hyperparameter combinations. Particularly NN and BRT, when tuned appropriately, showed surprisingly low bias for the estimated effects under feature collinearity (Fig. 1, S5, S6), which allows them to correctly adjust for confounding and other causal structures if the feature selection is causally constrained. This means that if causal connections between features and the response are known, ML algorithms and in particular NNs appear to be a viable alternative to statistical models for adjusting for confounders and estimating feature effects.

## Understanding the mechanism behind biased of feature effects under collinearity

The different susceptibility of the examined ML algorithms to bias induced by collinearity is presumably the result of different explicit and implicit algorithmic regularization mechanisms in these algorithms. For the elastic net, the regularization and thus the cause of the bias is explicit (and there is work to correct the models for the spillover bias). Also, for RF, it is relatively clear that the random subsampling of features creates an implicit regularization which explains the strong causal spillover observed in our simulations. For other algorithms such as BRT, we can only speculate about the mechanisms behind the observed biases: Our naive BRT implementation showed that pure boosting with linear models can be unbiased (Fig. S3), while boosting with regression trees (Fig. S3) can lead to either causal spillover or causal greediness (Fig. S3). In our simulations, state-of-the-art BRT implementations (used in Figs. 2, 3, 4) seem to prevent the causal greediness effect and only displayed causal spillover. Note that this is even though we specifically avoided "extreme boosting," which introduces boosting and dropout into BRT (42), which would likely cause additional spillover.

It was often reported that also NNs exhibit a so-called simplicity bias in their predictions with a negative impact on their generalizability, potentially caused by the stochastic gradient descent and not wide enough layers (24, 25, 43). A predictive simplicity bias should transfer to feature effects, suggesting that also NNs should exhibit causal spillover. We did not find such an effect for unregularized NNs, but we did find that with strong collinearity, both boosting and NN required far more boosting respectively optimization steps that what is needed to obtain reasonable predictive errors until they successfully separated the features (Fig. S4,

S13). Reported simplicity biases could thus also be explained by the common approach to stop training once the cross-validation loss does not further improve.

## Hyperparameters control bias-variance trade-off for effect estimates

For all algorithms, hyperparameters had substantial effects on the observed biases, especially in data-poor situations (Fig. 2). While some effects, for example the regularization parameters in the elastic net, were as expected, others like the choice of the activation function in the NN were surprising: SELU strongly reduced the bias of effect estimates and prediction errors (Fig. 2). The fact that this SELU effect diminished with increasing number of observations, that we used structured (tabular) data, or that we used a regression and not a classification task (44) may explain why this wasn't discovered before (Fig. S9).

Hyperparameters often had opposite effects on the bias and variance of effect estimates (and prediction errors) (Fig. 2), reflecting the expected trade-off between bias-variance when tuning regularization parameters. More importantly, however, the shape of this bias-variance tradeoffs differed for effect estimates versus prediction errors (e.g., mtry in RF), resulting in different sets of optimal hyperparameters (Fig. 2). This confirms the common expectation that there is a trade-off between tuning models for prediction and explanation. However, the difference between the two was not large, which is reassuring, given that in practical applications, hyperparameter tuning is only possible for the prediction error.

## Advantages and challenges when using ML models for inference

A question that remains is why and when we should prefer a causally constrained ML algorithms over OLS, which has the advantage of being the best linear unbiased estimator (BLUE). We believe in practice, there are two major drawbacks of OLS or other parametric regression models. First, an OLS requires that the model structure is specified a priori. If this structure is incorrectly specified, the effect of confounder, for example, may not correctly be adjusted, which can induce bias and causal spillover (cf. 44). Related to this, in practice, sample sizes are often prohibitively small for specifying a model with all possible effects, which that analysts either must make ad how decisions or accept that the variance of estimates his high and thus the power to see effects low. Machine learning approaches can potentially better trade-off bias against variance, and it is further possible to tune this trade-off to metrics that are particularly important for practitioners. For example, our trained ML models had a high reliability at identifying zero effects in the data-poor situations where the OLS failed to fit (Fig.

3). Among the ML and DL models, elastic-net showed the lowest errors, but we note that this was for a classical elastic net on top of an OLS where we prescribed linear effects. In this case, this worked well because we simulated data with linear effects, but we assume that in real-world scenarios NN will outperform elastic-net unless for the presence nonlinear effects or feature interactions happens to be guess exactly right (Fig. S1).

Comparing our approach to other causal ML algorithms, we see the closest resemblance to double / debiased ML, which uses a two-step process, where in a first step, two models are trained to predict the explanatory variable and the response based on the confounder (adjustment step), and then a final model is trained on the residuals of the first models to estimate the (adjusted) effect of the predictor (estimation step) (8). The validity of this approach was first proven for OLS (46) and depends only on the unbiasedness of the models involved (8). For OLS, the approach does not provide any advantage over a direct adjustment in a multiple regression for linear effects. For ML models, the advantage is that the adjustment and estimation models can be independently tuned and chosen. Our approach, on the other hand, which essentially generalizes a multiple regression model, seems to us easier to implement, understand, and may have advantages in particular predictive scenarios. Another alternative is the popular causal forest algorithm, which also essentially corrects for confounders while predicting the effect of a target feature. We view it as a task for further research to better understand the practical advantages and disadvantages of these alternative approaches. In particular, we believe it would be important to understand which of those approaches leads to a lower error on the estimated causal effects in situations that are representative for practical analysis in psychology, economics, medicine or ecology.

A limitation of all approaches discussed here, including OLS, is that they assume that the causal relationship between the characteristics is known a priori, so that our task is only to adjust for it. In practice, this assumption can often be met because the directions of the effects can be inferred from existing scientific knowledge, but when this is not the case, they must be estimated from the data, which is still extremely challenging.

## Advantages of causally constrained models for out-of-distribution predictions

Contradictory to the general assumptions that good predictive and explanatory models differ, we show that causal constraints that aid the model in learning the true underlying causal structure can also aid predictions when the collinearity structure of the feature space changes (out-of-distribution). This is not particularly surprising because it is well-known, even for statistical models, that selecting features causally is not necessarily beneficial for obtaining the lowest in-distribution prediction error, but it may help for out-of-distribution predictions where feature collinearity is changed (22)

Using a hypothetical example of predictions of lung cancer risk in an observational study and a clinical trial, we highlight that these effects could have important real-world applications. In our example, we find that a non-causal model has lower predictive in-distribution error, but higher out-of-distribution error compared to a causally constrained ML model (Fig. 4). We note that apart from the fact that the causally constrained model generalizes better, it has the additional advantage of being interpretable in terms of causal effects, which is of interest for science and clinical practitioners, but possibly also for questions of fairness in AI (47).

While it is generally understandable why certain algorithms (in particular RF) show higher biases on inferred feature effect under collinearity (see above), we wonder if these effects have any advantages for in-distribution predictions. It is interesting that state-of-the art BRT algorithms that often show the best performance on tabular data added algorithmic features similar to the random forest on the vanilla algorithm that has lower biases on the effects. We speculate that the spillover caused by the model averaging underlying these additions may actually be helpful in improving stability and reducing variance of the predictions, thus suggesting again that some algorithms may be better suited for in-distribution predictions, while others are better suited for inference or out-of-distribution predictions.

## Conclusion

Certain ML and DL algorithms, in particular neural networks, can approximately estimate the effect of one or several target features, adjusted for the effect of other features. Thus, these models can in principle be used like a multiple regression, and if needed, confidence intervals and p-values could be calculated on top based on bootstrapping. The observations that such causally constrained models may have larger in-distribution but lower out-of-distribution predictive errors, together with the fact that tuning hyperparameters for prediction is often a good proxy for inference as well suggests to us that the trade-off between predictive modelling and inference may not be as wide and deep as often assumed.

These results have significant implications for both predictive and explanatory modeling. For predictive modelling, they suggest that causally constraining ML and DL models can reduce out-of-distribution prediction error, which may often be a practically relevant objective. For explanatory modeling, it shows that ML algorithms such as BRT and NN can produce reliable inferences. Although more research is needed to better understand their biases and offer appropriate statistical guarantees on effect estimates, their higher flexibility provides at least the theoretical perspective that they could outperform traditional methods in situations with many nonlinearities or higher order interactions, which may actually account for the majority of applied statistical analyses of observational data.

# Methods

Statistical analysis and simulations were conducted in R (version 4.0.5, R Core Team, 2021). All code for reproducing our analysis can be found in https://github.com/MaximilianPi/Pichler-and-Hartig-Causal-ML. We additionally archive this code in persistent repository upon acceptance of the manuscript.

## Definition of average conditional effects

To extract the feature effects in a trained ML or DL model, we use average conditional effects (ACE), which are also known under the name average marginal effects. Consider a feature matrix $X = (x_1, ..., x_k)'$ with $k$ feature vectors and a response vector $y$ with their true relationship $y = f(X)$ and $\hat{f}(\cdot)$ is estimated by ML algorithms. Because the trained relationship can be highly complex, we find different conditional effects ($CE_{ik}$) (or interactions) for each observation $i$ of the $k$-th feature vector in the feature space. The $CE_k$ for feature vector $x_k$ is then $CE_k = \frac{\partial \hat{f}(X)}{\partial x_k}$ which is approximated by $CE_k \approx \frac{\hat{f}(x_1, x_2, ..., x_k+h, ..., x_j) - \hat{f}(x_1, x_2, ..., x_k, ..., x_j)}{h}$, $h > 0$. The conditional effects for $x_k$ ($CE_k$) are then averaged to $ACE_k$.

For linear effects, any average will produce an ACE that asymptotically corresponds to the coefficients in linear regression models ($y = \beta_1 x_1 + \cdots + \beta_k x_k, \beta_k \approx ACE_k$). For non-linear feature effects, the problems arise that dense areas in the feature space would be overrepresented in an arithmetic average. There have been several proposals how to average in such a case (34). As we did not consider nonlinear effects in our simulation, our results are not affected by this problem, but in general, we propose to average the $ACE_k = \sum_{i=1}^{N} w_i \, CE_{ik}$, with weights $w_i$ proportional to the inverse of the estimated density in the feature space of $x_k$.

## Near-asymptotic performance

We first simulated two different scenarios (Fig. 1, first column) with a large sample size of 1000 observations. This sample size is large enough so that effects of stochasticity induced by the data generation process and default hyperparameters for each model can be neglected. The two scenarios were a) a base scenario with three independent features, one without an effect, and b) a mediator scenario with two features forming a mediator path and a third feature independently affecting the response (for more details, see Methods). We fitted linear regression models (LM) to each scenario as a reference and compared the estimates to the effects learned by the ML models extracted by the ACE.

We simulated two scenarios with different collinearity structures. In all three scenarios we simulated five features $(x_1, x_2, x_3, x_4, x_5)$ and one response vector $y$. In the first scenario, the data generating model was $y \sim N(1.0 \cdot x_1 + 0.0 \cdot x_2 + 1.0 \cdot x_3 + 0.0 \cdot x_4 + 0.0 \cdot x_5, \sigma)$ with $\sigma = 0.3$ and all five features independent of each other (no collinearity). The feature matrix $X$ was sampled from a multivariate normal distribution with mean vector $\mu = 0$ and the covariance matrix being the identity. In the second scenario, the data generating model was the same but $\Sigma$ which was used to sample the feature matrix $X$ had an entry of 0.9 ($\Sigma_{1,2} = \Sigma_{2,1} = 0.9$) so that $x_1$ and $x_2$ were highly correlated. We sampled from each scenario 1000 observations.

Model fitting and evaluation

We fitted RF (Wright & Ziegler, 2017, 100 trees), BRT (Chen & Guestrin, 2016; 140 trees; "req:squarederror" objective function), NN (Amesöder & Pichler, 2022; three hidden layers with each 50 units; reLU activation functions; batch size of 100; AdaMax optimizer; learning rate of 0.01; 32 epochs), linear regression model (lm function), and glmnet (Friedman et al., 2010 and Ooi, 2021) packages; alpha = 0.2; lambda was tuned via 10-fold) to the data generated by the three scenarios (1,000 observations) ($X$ as feature matrix and $y$ as response vector). Afterwards, we calculated the individual ACE for each of the five features. We repeated the procedure (sampling from the scenarios and fitting the models to the data) 100 times and averaged the results.

As the simulated effects are linear, the theoretical ACE are equivalent to the true linear effects used in the data generating models: $\widehat{ACE_i} \approx \hat{\beta}_i$. To assess bias and variance, we calculated the bias $Bias = \beta_i - \widehat{ACE_i}$ and the variance of the $\widehat{ACE_i}$ over 500 replicates for all five features.

Performance in data-poor situations

We assume that we are interested in two effects, $\beta_1 = 1.0$ and $\beta_2 = 0.0$. The other effects were equally spaced between zero and 1.0. Features were sampled from a multivariate normal distribution with a covariance matrix ($\Sigma$) sampled from a LKJ distribution ($\eta = 2$) so that the features were weakly correlated on average. We calculated bias and the variance for the two target effects and all models.

The data generating model was $y \sim N(X\beta, \sigma)$ with $\sigma = 0.3$ with $X$ being the feature matrix (100 features) and $\beta$ the effect vector. $X$ was sampled from a multivariate normal distribution with mean vector $\mu = 0$ and distribution and the covariance matrix ($\Sigma$) was sampled from a LKJ distribution ($\eta = 2$) so that the features were weakly correlated on average. Effects were $\beta_1 = 1.0$ and $\beta_2 = 0.0$ and rest of the effects (98) were equally spaced between zero and 1.0.

Hyperparameter tuning

We performed a hyperparameter search to check if and how hyperparameters influence differently or equally effect estimates and the prediction error, so does a model tune after the prediction error has biased effects? For that, we created data-poor simulation scenarios with the above described data generating model and 50, 100, and 2000 observations and 100 features with effects ($\beta_i, i = 1,\ldots,100$), $\beta_1 = 1.0$, and $\beta_2$ to $\beta_3$ were equally spaced between 0.0 to 1.0 so that $\beta_2 = 0.0$ and $\beta_{100} = 1.0$.

Features were sampled from a multivariate normal distribution and all features were randomly correlated (Variance-covariance matrix $\Sigma$ was sampled from an LKJ-distribution with $\eta = 2.0$.

1,000 combinations of hyper-parameters were randomly drawn (Table S1). For each draw of hyperparameters, the data simulation and model fitting were repeated 20 times. $\widehat{ACE}_1$ and $\widehat{ACE}_2$ were recorded (for each hyperparameter combination and for each repetition). Bias, variance, and mean square error (MSE) were calculated for estimated effects and the average (over the 20 repetition) MSE for predictions on a holdout of the same size as the training data.

To understand how hyperparameters affect bias, variance, and MSE of estimated effects and predictions, we fitted generalized additive model (GAM) on the hyperparameters with the respective errors as response. The average responses were first subtracted from the responses to set the intercept to 0 (we suppressed the intercept in the GAMs because we were not interested in a reference level). We also fitted a random forest (2000 trees to get stable effects) on the hyperparameters to get variable importances for all hyperparameters.

To get the optimal hyperparameters, we fitted random forest models on the hyperparameters of the MSE for the estimated effect $\hat{\beta}_1$ and the predictions $\hat{y}$. We then predicted for all hyperparameters and selected the hyperparameters with the lowest MSE (Table S2, Table S3).

Model fitting and evaluation

We fitted RF (49), BRT (42), NN (50), linear regression model (lm function), and elastic-net (51, 52) to the data generated for 50, 100, and 600 observations.

We calculated the individual ACE for the first two effects $\beta_1 = 1.0$ and $\beta_2 = 0.0$. We repeated the procedure (sampling from the scenarios and fitting the models to the data) including the sampling of the covariance matrix $\Sigma$ 1000 times. We calculated bias and variance for both effects.

## Acknowledgements

We thank Tankred Ott for valuable comments on the manuscript.

## Author Contributions

MP and FH jointly conceived and designed the study. Both authors contributed equally to the writing and preparation of the manuscript.

## Data availability

Code to reproduce the analysis can be found in the following repository https://doi.org/10.5281/zenodo.8052354.

# Supporting Information

## 1 Extending ACE to two-way interactions

ACE can be extended to $n$-dimensions to detect $n$ way predictor interactions. Here, we extended ACEs to two dimensions to detect two-way predictor interactions by asking what the change is of $\hat{f}(\cdot)$ when predictors $\boldsymbol{x}_m$ and $\boldsymbol{x}_k$ change together:

$$\mathbf{CE}_{mk} = \frac{\partial^2 \hat{f}(\mathbf{X})}{\partial \boldsymbol{x}_m \, \partial \boldsymbol{x}_k}$$

We can approximate $\mathbf{CE}_{mk}$ with the finite difference method:

$$\mathbf{CE}_{mk} \approx \frac{\hat{f}(\boldsymbol{x}_1, \boldsymbol{x}_2, \ldots, \boldsymbol{x}_m + h, \boldsymbol{x}_k + h, \ldots, \boldsymbol{x}_j)}{2(h_m + h_k)} - \frac{\hat{f}(\boldsymbol{x}_1, \boldsymbol{x}_2, \ldots, \boldsymbol{x}_m - h, \boldsymbol{x}_k + h, \ldots, \boldsymbol{x}_j)}{2(h_m + h_k)}$$
$$- \frac{\hat{f}(\boldsymbol{x}_1, \boldsymbol{x}_2, \ldots, \boldsymbol{x}_m + h, \boldsymbol{x}_k - h, \ldots, \boldsymbol{x}_j)}{2(h_m + h_k)}$$
$$- \frac{\hat{f}(\boldsymbol{x}_1, \boldsymbol{x}_2, \ldots, \boldsymbol{x}_m - h, \boldsymbol{x}_k - h, \ldots, \boldsymbol{x}_j)}{2(h_m + h_k)}$$

$h_m$ and $h_k$ are set to $0.1 \cdot sd(\boldsymbol{x}_m)$ and $0.1 \cdot sd(\boldsymbol{x}_k)$. All predictors are centered and standardized.

### 1.1 Proof of concept simulations for inferring interactions

To test the ability of ML algorithms to identify predictor-predictor interactions, we repeated the proof-of-concept simulations, but with an interaction between $X_1$ and $X_2$. The data generation model was $Y \sim 1.0 \cdot X_1 + 1.0 \cdot X_5 + 1.0 \cdot (X_1 \cdot X_2) + \epsilon$ with $\epsilon \sim N(0, 1.0)$. We simulated two scenarios, in the first ("collinear") $X_1$ and $X_2$ were collinear (Pearson correlation factor = 0.9) and in the second without collinearity between the predictors.

We sampled 1000 and 5000 observations from each scenario. The ML algorithms (RF, BRT, NN, and NN with dropout) were fit to the data without predictor engineering the predictor interactions (because ML algorithms are known to be able to infer interactions automatically), while the regression algorithms (LM, l1, l2, and elastic-net) received all combinatorially possible predictor interactions as possible predictors. All effects were inferred using ACE. The bias was calculated for the interaction $x_1:x_2$.

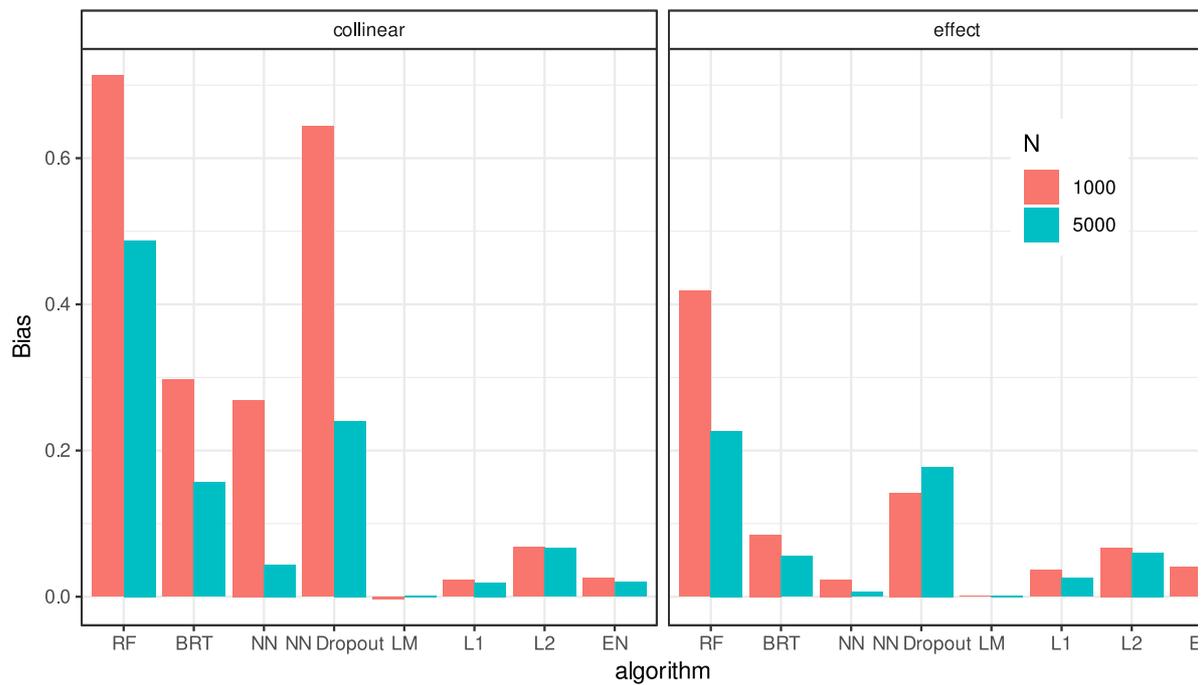

**Figure S** 1: Bias of proof-of-concept simulations in inferring two-way interactions between predictors. First panel shows results for simulations (200 repititions) for 1000 and 5000 observations with collinear predictors (Pearson correlation factor = 0.9 between $\mathbf{x}_1$ and $\mathbf{x}_2$). Second panel shows results for simulations (200 repititions) for 1000 and 5000 observations with without collinear. Red bars correspond to 1000 observations and blue bars to 5000 observations.

We found that for the ML algorithms (RF, BRT, and NN) NN showed the lowest for all scenarios (Fig. S1). Also collinearity increased the bias for the ML algorithms. No collinearity or more observations decreased the bias (Fig. S1). The regression models, LM, LASSO and Ridge regression, and elastic-net showed the lowest and in case of LM, no bias. However, we want to note here that the regression models received all possible predictor-predictor interactions as predictors while the ML algorithms had to infer the interactions on their own. Whit this in mind, the performance of the NN is surprising well, even competing with the penalized regression models. On the other hand, NN with dropout showed larger biases than BRT (Fig. S1).

## 1.2 Weighted ACE

If the instances of a predictor $x_j$ are not uniformly distributed, we propose to calculate a weighted $wACE_k = \Sigma_{i=1}^{N} w_i ACE_{ik}$ with the $w_i$ being, for example, the inverse probabilities of an estimated density function over the predictor space of $x_k$.

To demonstrate the idea of weighted ACE, we simulated a scenario with one predictor where the $\beta_1 = 2$ for values of the predictor $< 2$ and for the other predictor values $\beta_1 = 0$ (Fig. S2). The predictor was sampled from a log-Normal distribution. We fitted a linear regression model and a NN on the data and compared the effect estimated by the LM, the unweighted ACE, and the weighted ACE.

The LM estimated an effect of 1.48, the unweighted ACE was 1.95, and the weighted ACE was 1.48 (Fig. S2).

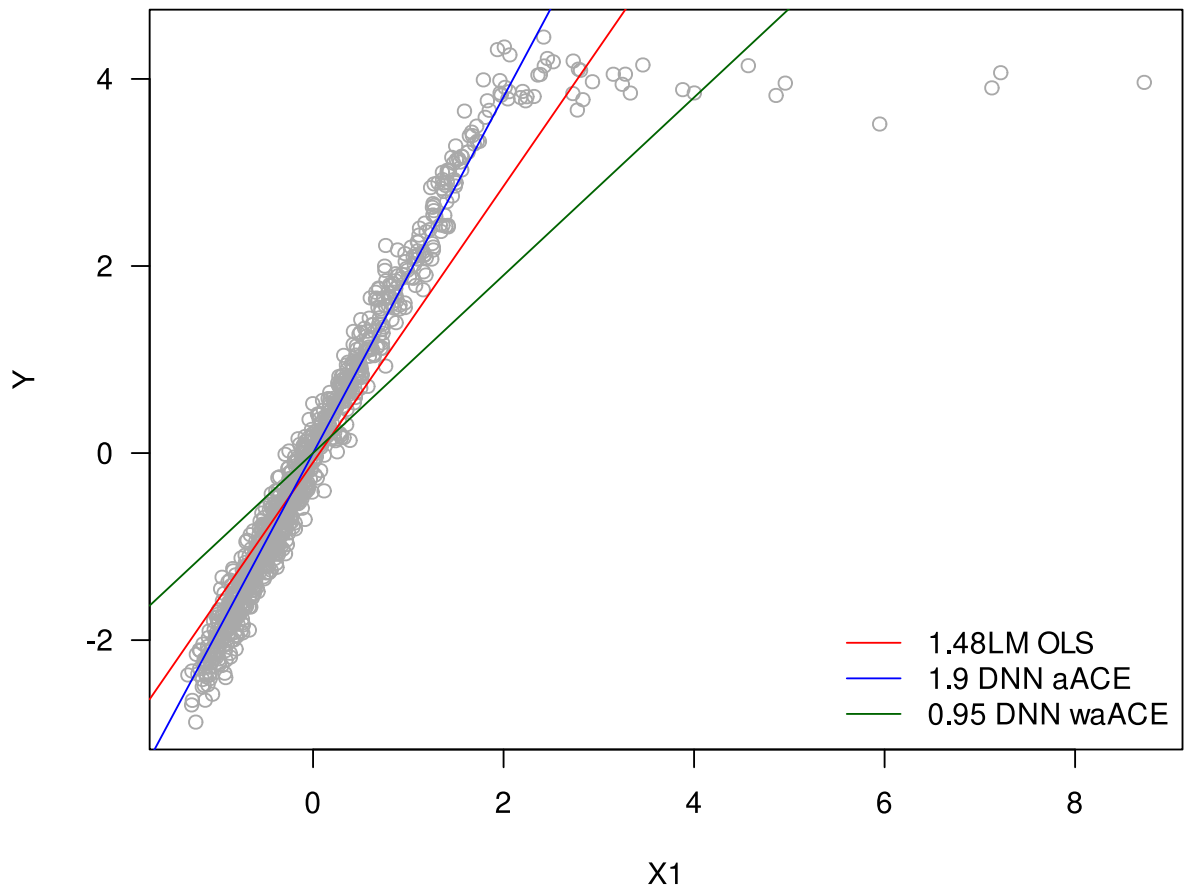

**Figure S** 2: Simulation example with non-uniform sampled predictor $x_1$ (log normal distributed). The red line is the effect estimated by a LM OLS. The blue line is the effect reported by an unweighted ACE from a NN. The green line is the effect reported by a weighted ACE from a NN.

## 2 Boosting and regression trees

### 2.1 Unbiasedness

Random forest (RF) and boosted regression trees (BRT) showed biased effect estimates in both scenarios, with and without collinearity, raising the question of whether the bias is caused by the boosting/bagging or the regression trees themselves. For RF, we know that the observed spillover effect is caused by the random subsampling (mtry parameter) in the algorithm, which explains the bias.

For BRT, however, it is unclear what is causing the bias (boosting or regression trees) because each member in the ensemble is always presented with all predictors (at least with the default hyperparameters, the BRT implementation in xgboost has options to use bootstrap samples for each tree and also subsamples of columns in each tree (or node), see Chen and Guestrin (2016)).

To understand how boosting and regression trees affect effect estimates, we simulated three different scenarios (Fig. S3, first column) without collinearity (Fig. S3a) and with collinearity (Fig. S3a, b) (we sampled 1000 observations from each data generating model (Fig. S3, first column) and estimated effects using ACE (500 repititions)).

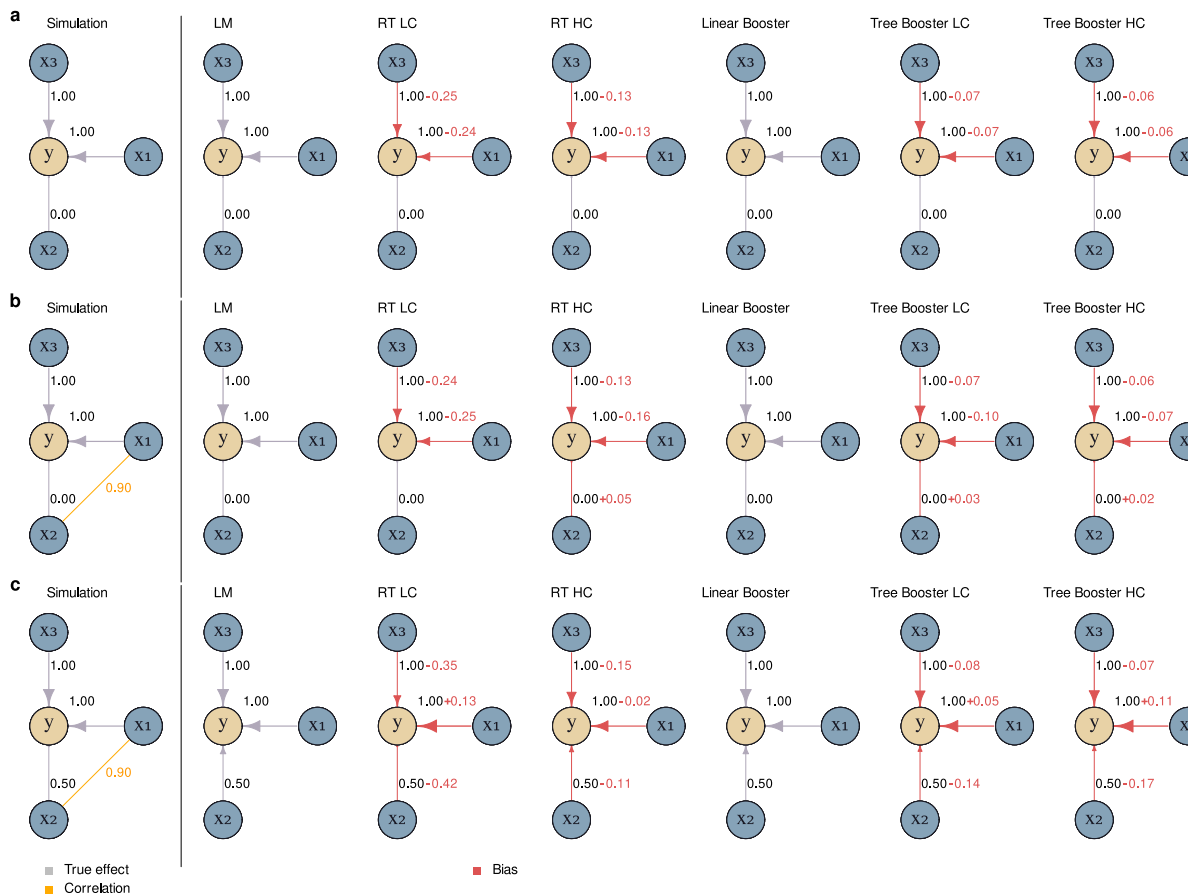

**Figure S** 3: Bias on effect estimates for different ML algorithms (LM = liner regression model (OLS), RT LC = regression tree with low complexity (depth), RT HC = regression tree with high complexity, Linear Booster, Tree Booster LC = tree booster with low complexity, Tree Booster HC = tree boster with high complexity) in three different simulated causal scenarios (a, b, and c). Sample sizes are so large that stochastic effects can be excluded (1000 observations). Effects of the ML models were inferred using average conditional effects. Row a) shows results for simulations with uncorrelated predictors with the true effect sizes. Row b) shows the results for simulations with $x_1$ and $x_2$ being strongly correlated (Pearson correlation factor = 0.9) but only $x_1$ has an effect on y (mediator) and row c) shows the results for $x_1$ and $x_2$ being strongly correlated (Pearson correlation factor = 0.9) with $x_1$ and $x_2$ having effects on y (confounder scenario).

We found that the regression tree (RT) is unable to estimate unbiased effects (Fig. S3), regardless of the presence or absence of collinearity or the complexity of the RT (depth of the regression trees). Without collinearity, effects in regression trees were biased toward zero, less so with higher complexity (Fig. S3). With collinearity, there was a small spillover effect for the RT with high complexity (Fig. S3b) to the collinear zero effect ($\beta_2$), similar to an l2 regularization. When the collinear predictor ($\beta_2$) had an effect (Fig. S3c), we found a stronger absolute bias for the smaller of the two collinear effects ($\beta_2$), confirming our expectation that RTs show a greedy effect. This greedy behavior was particularly strong for the low complexity RT (Fig. S3c).

To answer the question of how boosting affects the greediness and spillover effects of RT, we first investigated the behavior of a linear booster because of the well-known behavior of OLS under collinearity. And indeed, we found that the linear booster was unbiased in all three scenarios (compare LM and linear booster in Fig. S3), showing that boosting itself can produce unbiased effects.

Now, comparing the vanilla BRTs with low and high complexity (depth of individual trees) with the linear booster and the RTs, we found similar biases as for the RTs, in terms of spillover with a collinear zero effect and the greediness effect in the presence of a weaker collinear effect (Fig. S3).

## 2.2 Understanding boosting

Intuitive boosting shouldn't work because it's basically a regression of residuals. That is, and in the case of collinearity, the stronger of two collinear predictors in the first model would absorb the effect of the weaker second predictor that, for example, causes the omitted variable bias (the effect of the missing confounder is absorbed by the collinear effect).

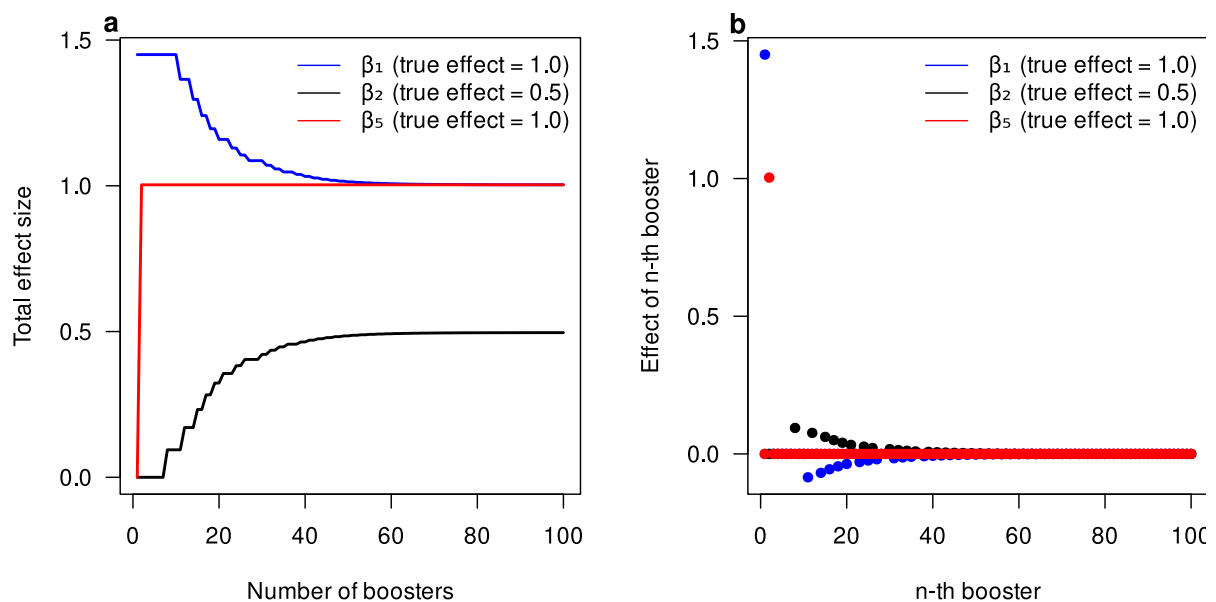

**Figure S** 4: Changes of effects within boosting. (A) shows the total effect of ensemble (linear booster) until the n-th ensemble member. (B) shows the effects of the n-th ensemble member. X1 and X2 were correlated (Pearson correlationf factor = 0.9).

Looking at the development of the total effect within a linear booster model (Fig. S4a), we found that the first members of the ensemble absorb the effect of the collinear effect ($\beta_1$ absorbed $\beta_2$, Fig. S4a), but as members are added to the ensemble, the collinear effect $\beta_2$ slowly recovers the effect of the stronger collinear effect until both are at their correct effect estimate (Fig. S4a). This retrieval works by reversing the sign of each member's effect, so that $\beta_1$, which initially has an effect of 1.5 (because it absorbed the effect of $\beta_2$), has small negative effects in subsequent trees, while $\beta_2$, which is initially estimated at 0, has small positive effects (Fig. S4b).

# 3 Proof of concept - Additional results

## 3.1 Addtional scenarios

To better understand the ability of ML algorithms in learning unbiased effects, we tested additional scenarios (Fig. S5, first column).

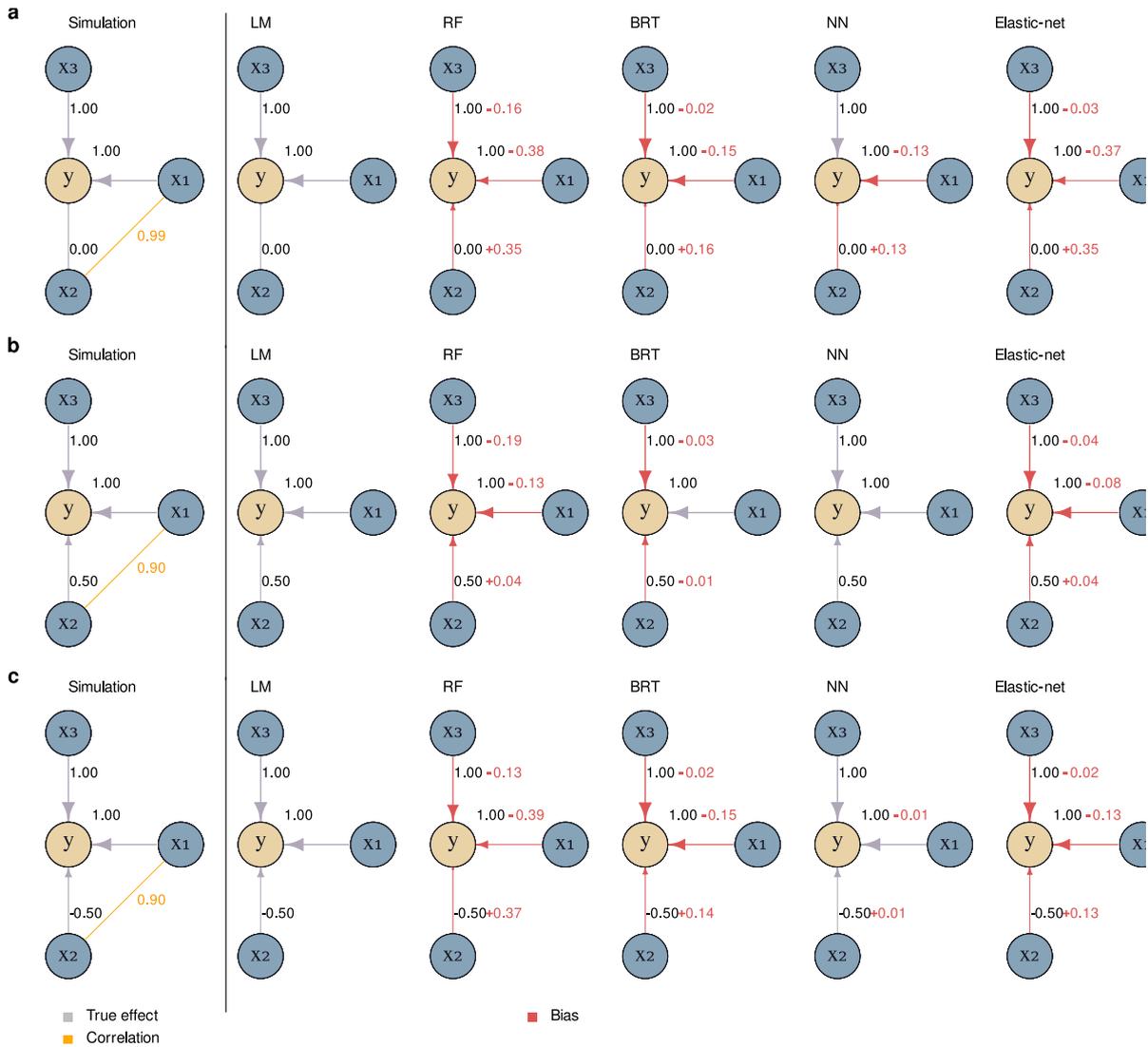

**Figure S** 5: Bias on effect estimates for different ML algorithms in trhee different simulated causal simulations (a, b, and c). Sample sizes are so large that stochastic effects can be excluded (1000 observations). Effects of the ML models were inferred using average conditional effects. Row a) shows the results for simulations with $X_1$ and $X_2$ being strongly correlated (Pearson correlation factor = 0.99) but only $X_1$ has an effect on y. Row b) shows results for simulations with with predictors (Pearson correlation factor = 0.5) with effect sizes ($X_1$: 1.0, $X_2$: 0.5, $X_3$: 1.0) and row c) shows results for simulations with with predictors (Pearson correlation factor = 0.5) with effect sizes ($X_1$: 1.0, $X_2$: -0.5, $X_3$: 1.0)

We found that NN cannot separate extreme collinear effects as the OLS (Fig. S5a) which, however, may improve with additional observations.

## 3.2 Additional models

To understand the different effects of regularization in NN (dropout), LASSO regression, and Ridge regression, we tested these models on our theoretical scenarios (Fig. S6, first column).

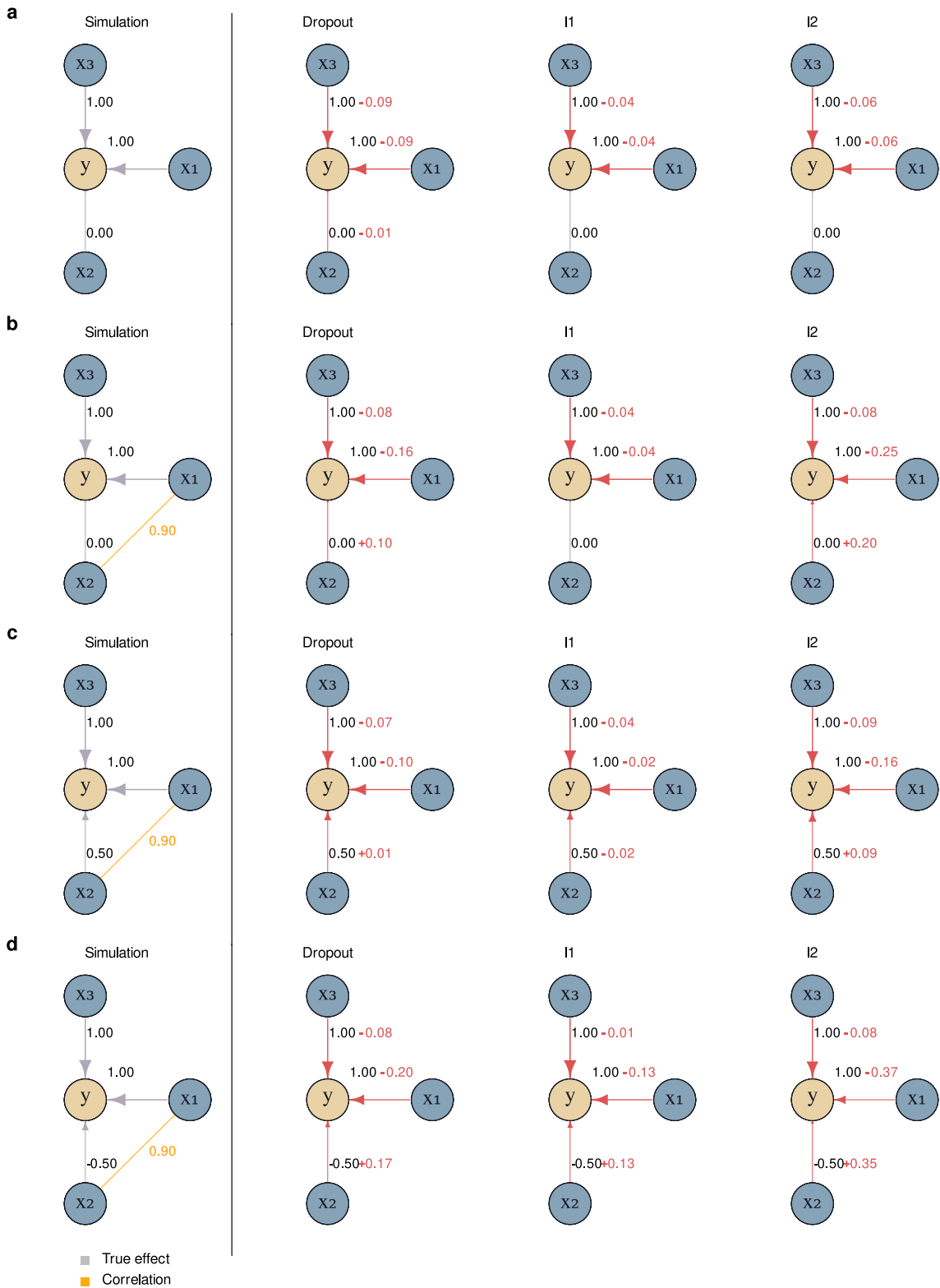

**Figure S 6**: Bias on effect estimates for different ML algorithms in two different simulated causal simulations (a and b). Sample sizes are so large that stochastic effects can be excluded. Effects of the ML models were inferred using average conditional effects. Row a) shows results for simulations with with predictors (Pearson correlation factor = 0.5) with effect sizes ($X_1$: 1.0, $X_2$: -0.5, $X_3$: 1.0). Row b) shows the results for simulations with $X_1$ and $X_2$ being strongly correlated (Pearson correlation factor = 0.99) but only $X_1$ has an effect on y.

Dropout has a negative effect on the ability to separate collinear effects in NN (Fig. S6) while also LASSO and Ridge (as expected) affect negatively the ability to separate collinear effects (Fig. S6).

# 4 Hyperparameter tuning

We performed a hyperparameter search to check if and how hyperparameters influence differently or equally effect estimates and the prediction error, so does a model tune after the prediction error has biased effects? For that, we created simulation scenarios with 50, 100, 600, and 2000 observations and 100 predictors with effects ($\beta_i, i = 1, \ldots, 100$) $\beta_1 = 1.0$, and $\beta_2$ to $\beta_3$ were equally spaced between 0.0 to 1.0 so that $\beta_2 = 0.0$ and $\beta_{100} = 1.0$.

Predictors were sampled from a multivariate normal distribution and all predictors were randomly correlated (Variance-covariance matrix $\Sigma$ was sampled from a LKJ-distribution with $\eta = 2.0$.

1,000 combinations of hyper-parameters were randomly drawn (Table S1). For each draw of hyperparameters, the data simulation and model fitting was repeated 20 times. Effect sizes of $X_1$ and $X_2$ were recorded (for each hyperparameter combination and for each reptition). Moreover, bias, variance, and mean square error (MSE) were recorded for the predictions on a holdout of the same size as the training data.

**Table S** 1: Overview over hyper-parameters for Neural Network, Boosted Regression Tree, and Random Forest

| Algorithm | Hyper-parameter | Range |
|---|---|---|
| Neural Network | activation function | [relu, leaky_relu, tanh, selu, elu, celu, gelu] |
| | depth | [1, 8] |
| | width | [2, 50] |
| | batch size (sgd) | [1, 100] in percent |
| | lambda | [2.65e-05, 0.16] |

| Algorithm | Hyper-parameter | Range |
| --- | --- | --- |
| Boosted Regression Tree | alpha | [0, 1.0] |
| | eta | [0.01, 0.4] |
| | max depth | [2, 25] |
| | subsample | [0.5, 1] |
| | max tree | [30, 125] |
| | lambda | [1, 20] |
| Random Forest | mtry | [0, 1] in percent |
| | min node size | [2, 70] |
| | max depth | [2, 50] |
| | regularization factor | [0, 1] |
| Elastic net | alpha | [0, 1.0] |
| | lambda | [0, 1.0] |

## 4.1 Results hyperparameter tuning

As described in the main text, we analyzed the effects of the hyperparameters on the different errors using GAMs and variable importance of random forest (Fig. S7, S8, S9).

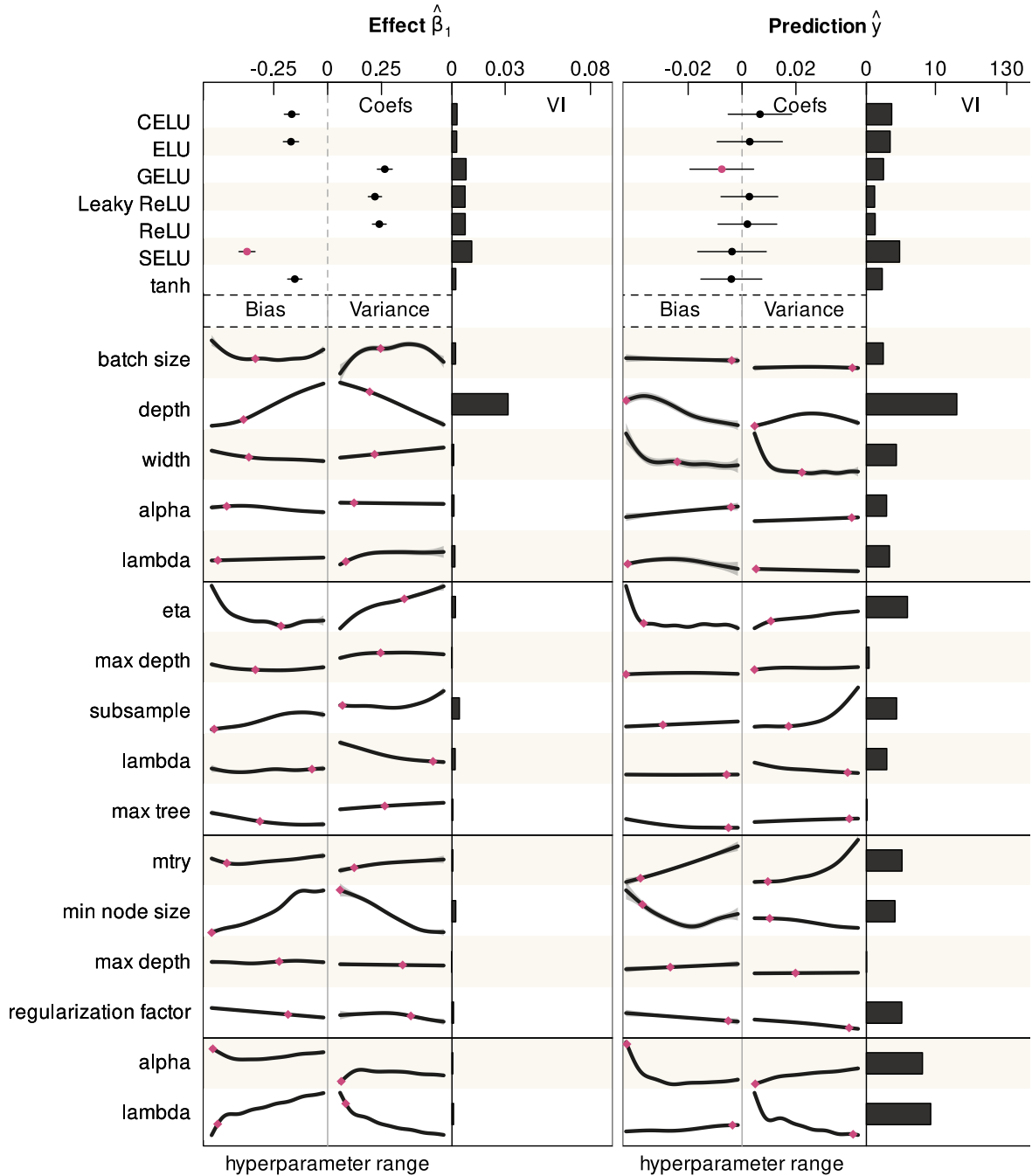

**Figure S** 7: Results of hyperparameter tuning for Neural Networks (NN), Boosted Regression Trees (BRT), Random Forests (RF), and Elastic Net (EN) for 50 observations with 100 predictors. The influence of the hyperparameters on effect $\hat{\beta}_1$ (bias, variance, and MSE)(true simulated effect $\beta_1 = 1.0$ ) and the predictions, $\hat{y}$ of the model (bias, variance, and MSE) were estimated by a multivariate generalized additive model (GAM). Categorical hyperparameters (activation function in NN) were estimated as fixed effects. The responses (bias, variance, MSE) were centered so that the categorical hyperparameters correspond to the intercepts. The variable importance of the hyperparameters was estimated by a random forest with the MSE of the effect $\hat{\beta}_1$ (first plot) or the prediction $\hat{y}$ (second plot) as the response. Red dots correspond to the best predicted set of hyperparameters (based on a random forest), in the first plot for the minimum MSE of the effect $\hat{\beta}_1$ and in the second plot for the minimum MSE of the predictions $\hat{y}$.

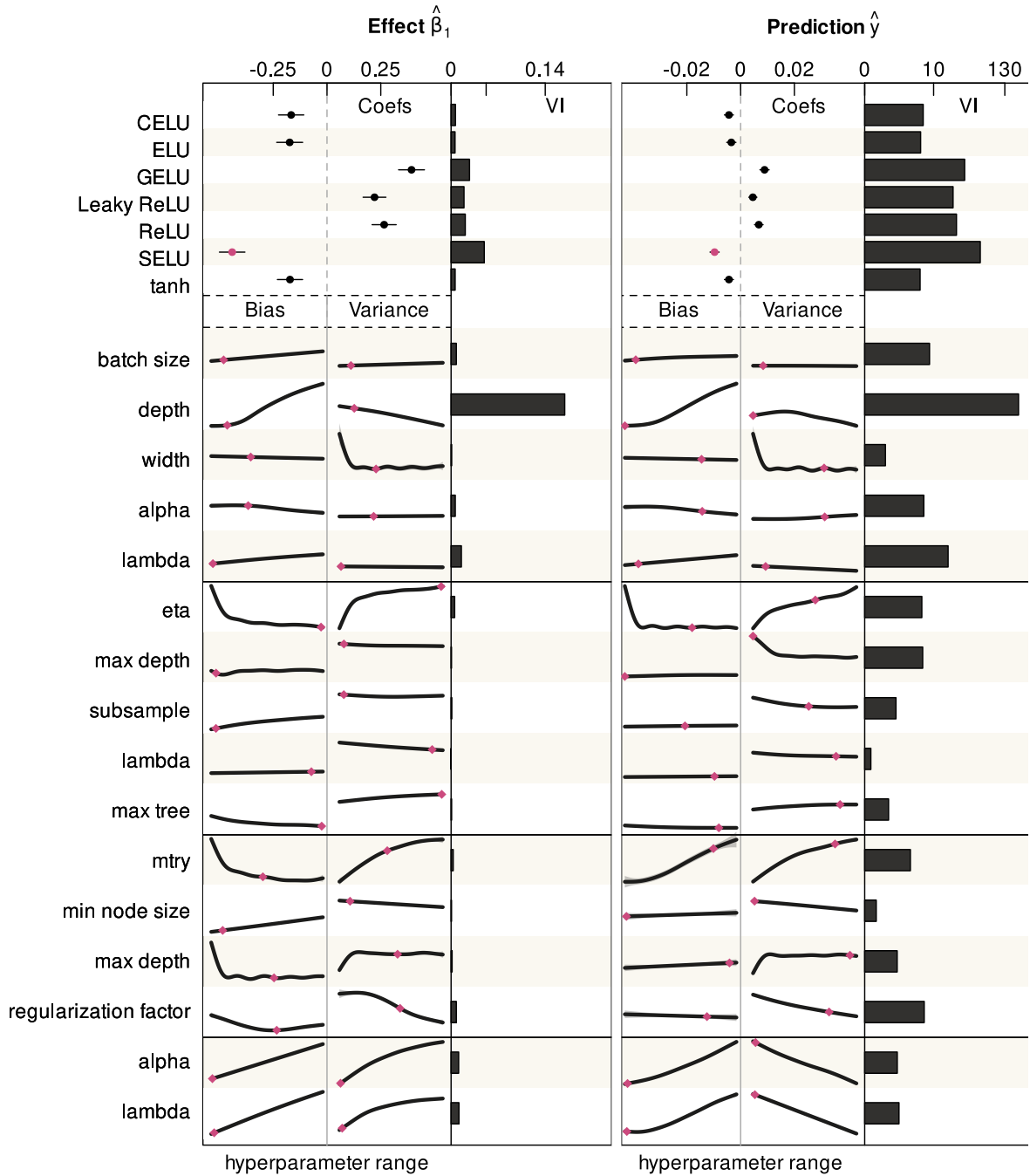

**Figure S 8**: Results of hyperparameter tuning for Neural Networks (NN), Boosted Regression Trees (BRT), Random Forests (RF), and Elastic Net (EN) for 600 observations with 100 predictors. The influence of the hyperparameters on effect $\hat{\beta}_1$ (bias, variance, and MSE)(true simulated effect $\beta_1 = 1.0$) and the predictions, $\hat{y}$ of the model (bias, variance, and MSE) were estimated by a multivariate

generalized additive model (GAM). Categorical hyperparameters (activation function in NN) were estimated as fixed effects. The responses (bias, variance, MSE) were centered so that the categorical hyperparameters correspond to the intercepts. The variable importance of the hyperparameters was estimated by a random forest with the MSE of the effect $\hat{\beta}_1$ (first plot) or the prediction $\hat{y}$ (second plot) as the response. Red dots correspond to the best predicted set of hyperparameters (based on a random forest), in the first plot for the minimum MSE of the effect $\hat{\beta}_1$ and in the second plot for the minimum MSE of the predictions $\hat{y}$.

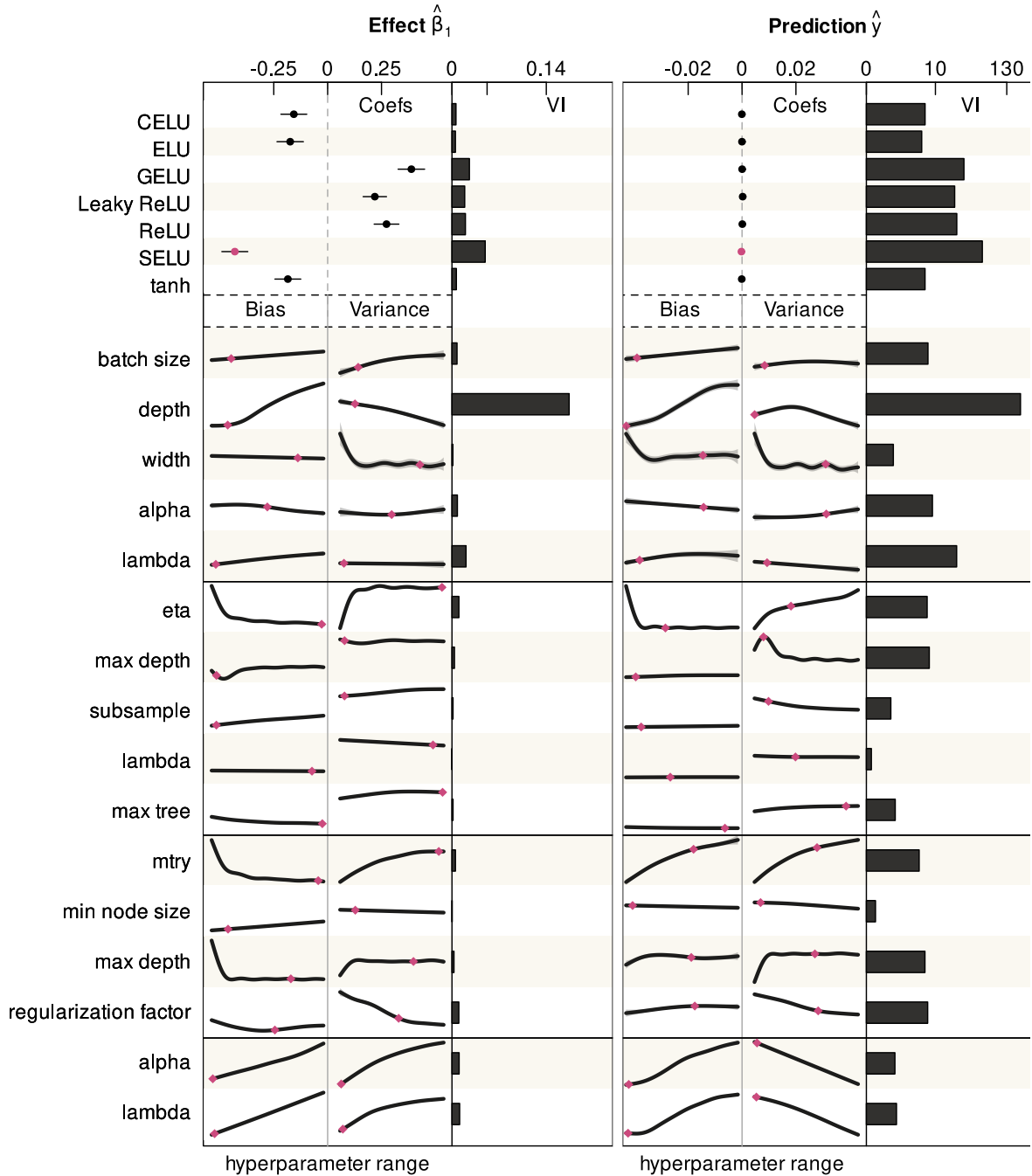

**Figure S** 9: Results of hyperparameter tuning for Neural Networks (NN), Boosted Regression Trees (BRT), Random Forests (RF), and Elastic Net (EN) for 2000 observations with 100 predictors. The influence of the hyperparameters on effect $\hat{\beta}_1$ (bias, variance, and MSE)(true simulated effect $\beta_1 = 1.0$ ) and the predictions, $\hat{y}$ of the model (bias, variance, and MSE) were estimated by a multivariate generalized additive model (GAM). Categorical hyperparameters (activation function in NN) were estimated as fixed effects. The responses (bias, variance, MSE) were centered so that the categorical hyperparameters correspond to the intercepts. The variable importance of the hyperparameters was estimated by a random forest with the MSE of the effect $\hat{\beta}_1$ (first plot) or the prediction $\hat{y}$ (second plot) as the response. Red dots correspond to the best predicted set of hyperparameters (based on a random

forest), in the first plot for the minimum MSE of the effect $\hat{\beta}_1$ and in the second plot for the minimum MSE of the predictions $\hat{y}$.

## 4.2 Optimal hyperparameters

The hyperparameters were chosen based on the lowest MSE for the predictive performance of the models (Table S2) and the lowest MSE for the effect ($\beta_1$) on $X_1$ (Table S3). The selection of the best hyperparameters was done by first fitting a random forest (default parameters) with the MSE as response and the hyperparameters as predictors, and then using the set of hyperparameters that predicted the lowest MSE.

**Table S 2:** Best predicted set of hyperparameter for ML algorithms (tuned after MSE of predictions)

| Algorithm | Hyperparameter | n = 50 | n = 100 | n = 600 | n = 2000 |
|---|---|---|---|---|---|
| NN | activations | celu | selu | selu | selu |
|  | sgd | 0.944 | 0.348 | 0.098 | 0.098 |
|  | depth | 1 | 1 | 1 | 1 |
|  | width | 24 | 20 | 35 | 35 |
|  | alpha | 0.939 | 0.821 | 0.693 | 0.693 |
|  | lambda | 0.003 | 0.02 | 0.019 | 0.019 |
| BRT | eta | 0.072 | 0.126 | 0.245 | 0.147 |
|  | max_depth | 2 | 2 | 2 | 4 |
|  | subsample | 0.666 | 0.511 | 0.77 | 0.57 |
|  | lambda | 9.073 | 8.888 | 8.21 | 4.556 |
|  | max_tree | 117 | 109 | 110 | 114 |
| RF | mtry | 0.129 | 0.466 | 0.792 | 0.603 |
|  | min.node.size | 12 | 2 | 3 | 6 |
|  | max.depth | 21 | 19 | 47 | 30 |
|  | regularization.factor | 0.914 | 0.874 | 0.736 | 0.615 |
| EN | alpha | 0.007 | 0.008 | 0.025 | 0.025 |
|  | lambda | 0.286 | 0.028 | 0.006 | 0.006 |

Table S 3: Best predicted set of hyperparameterfor ML algorithms (tuned after MSE of effect $X_1$)

| Algorithm | Hyperparameter | n = 50 | n = 100 | n = 600 | n = 2000 |
|---|---|---|---|---|---|
| NN | activations | selu | selu | selu | selu |
|  | sgd | 0.391 | 0.395 | 0.112 | 0.175 |
|  | depth | 3 | 3 | 2 | 2 |
|  | width | 18 | 40 | 19 | 39 |
|  | alpha | 0.135 | 0.613 | 0.332 | 0.498 |
|  | lambda | 0.009 | 0.011 | 0.002 | 0.006 |
| BRT | eta | 0.252 | 0.327 | 0.393 | 0.393 |
|  | max_depth | 11 | 17 | 3 | 3 |
|  | subsample | 0.514 | 0.584 | 0.523 | 0.523 |
|  | lambda | 9.051 | 7.779 | 9.053 | 9.053 |
|  | max_tree | 71 | 102 | 124 | 124 |
| RF | mtry | 0.137 | 0.926 | 0.462 | 0.952 |
|  | min.node.size | 2 | 4 | 9 | 12 |
|  | max.depth | 31 | 29 | 29 | 36 |
|  | regularization.factor | 0.683 | 0.894 | 0.587 | 0.566 |
| EN | alpha | 0.011 | 0 | 0.011 | 0.011 |
|  | lambda | 0.016 | 0.018 | 0.009 | 0.009 |

# 5 Additional results for data-poor scenarios

## 5.1 Prediction error of scenarios

Fig. S10 shows the MSE of the predictions on the holdouts for the different ML algorithms and different number of observations of the data-poor scenarios (see main text).

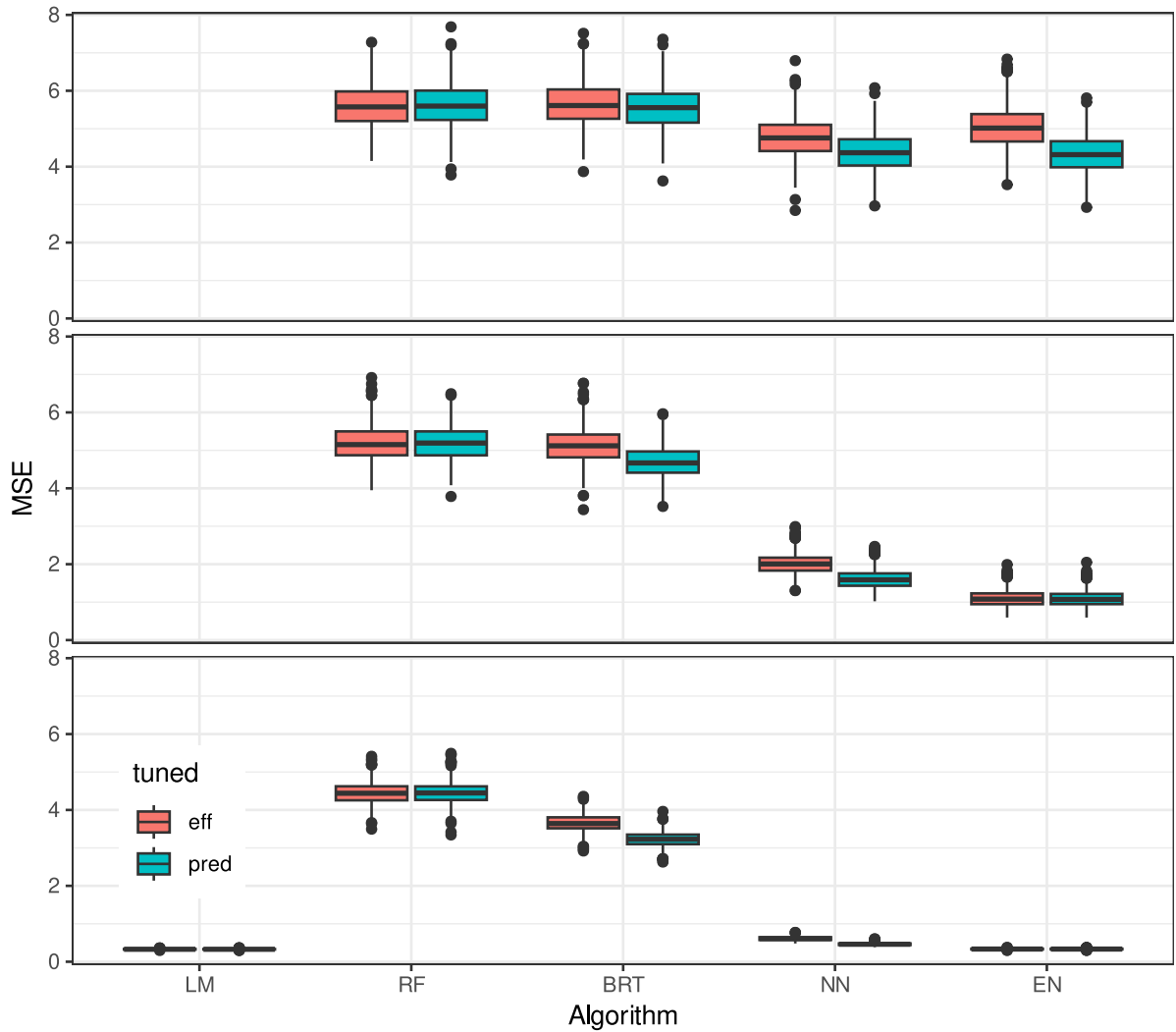

**Figure S** 10: Prediction error (mean square error, MSE) of data poor simulations with optimal hyperparameters either tuned after the best MSE of the effect size (red) or the best MSE of the prediction error (blue).

# 6 Data-poor scenarios without collinearity

## 6.1 Bias and variance of effects

To assess the effect of collinearity on the data-poor simulations, we repeated the scenarios but without collinearity. $\Sigma$ which was used in the sampling process of the predictor matrix (multivariate normal distribution) was set to the identity matrix. While it is not ideal, we used the best hyperparameters (Table S3, Table S4) which were tuned for the collinear scenarios, for these scenarios

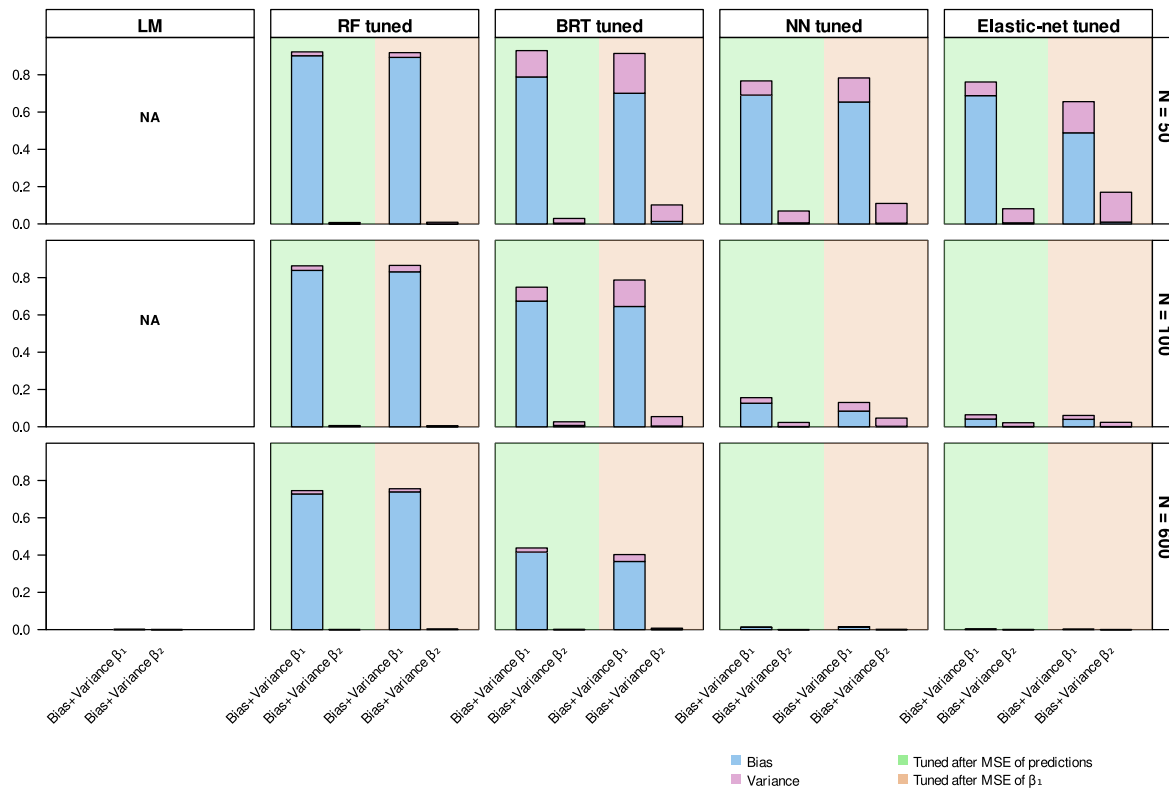

**Figure S 11**: Bias and variance of estimated effects in data-poor situations. N = 50, 100, and 600 observations of 100 uncorrelated predictors were simulated. True effects in the data generating model were $\beta_1$=1.0, $\beta_2$=0.0, and the other 98 effects were equally spaced between 0 and 1. Models were fitted to the simulated data (1000 replicates) with the optimal hyperparameters (except for LM, which doesn't have hyperparameters). Hyperparameters were selected based on the minimum MSE of ($\hat{\beta}_1$) (green) or the prediction error (based on $\hat{y}$) (red). Bias and variance were calculated for $\hat{\beta}_1$ and $\hat{\beta}_2$. Effects $\hat{\beta}_i$ for $i = 1, \ldots, 100$) were approximated using ACE.

We found similar results as for data-poor scenarios with collinearity (Fig. S11). NN and elastic-net show the lowest errors and strongest increase in those errors with increasing number of observations (Fig. S11).

## 6.2 Prediction error of scenarios

Fig. S12 shows the prediction errors for the ML algorithms for the data-poor simulations without collinearity. We found similar results as for the data-poor simulations with collinearity.

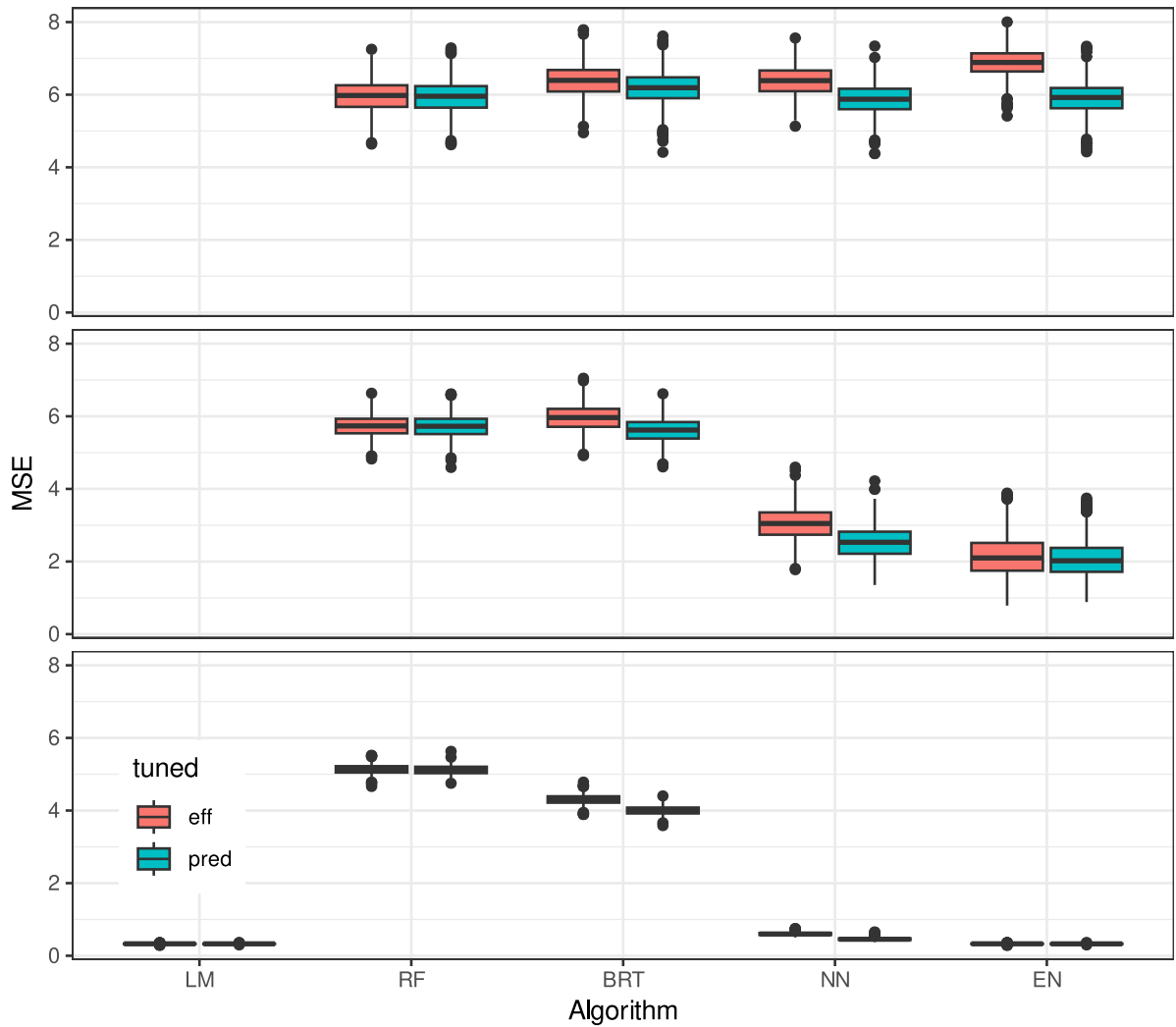

**Figure S**12: Prediction error (mean square error, MSE) of data poor simulations with optimal hyperparameters either tuned after the best MSE of the effect size (red) or the best MSE of the prediction error (blue).

## 7 Learning in neural networks

To understand the internal learning of neural networks, we trained neural networks of two different sizes (3 layers of 50 units and 3 layers of 500 units) on a simple collinear scenario ($Y \sim 1.0 \cdot X_1 + 0.0 \cdot X_2 + \epsilon, \epsilon \sim N(0,0.3)$; $X_1$ and $X_2$ were collinear (Pearson correlation factor = 0.9)) and calculated the ACE after each batch optimization step.

We found that the estimates of the botch effect were initially estimated to be around 0 (Fig. S13 A, B), probably due to the initialization of the neural networks, which resembles a shrinkage behavior (weights have to be moved away from 0 step by step in the gradient descent). After this initialization phase, both estimates are within the expected negative log-likelihood surface of OLS (Fig. S13C) and are estimated over the training period to the correct estimates ($X_1 = 1.0$ and $X_2 = 0.0$).

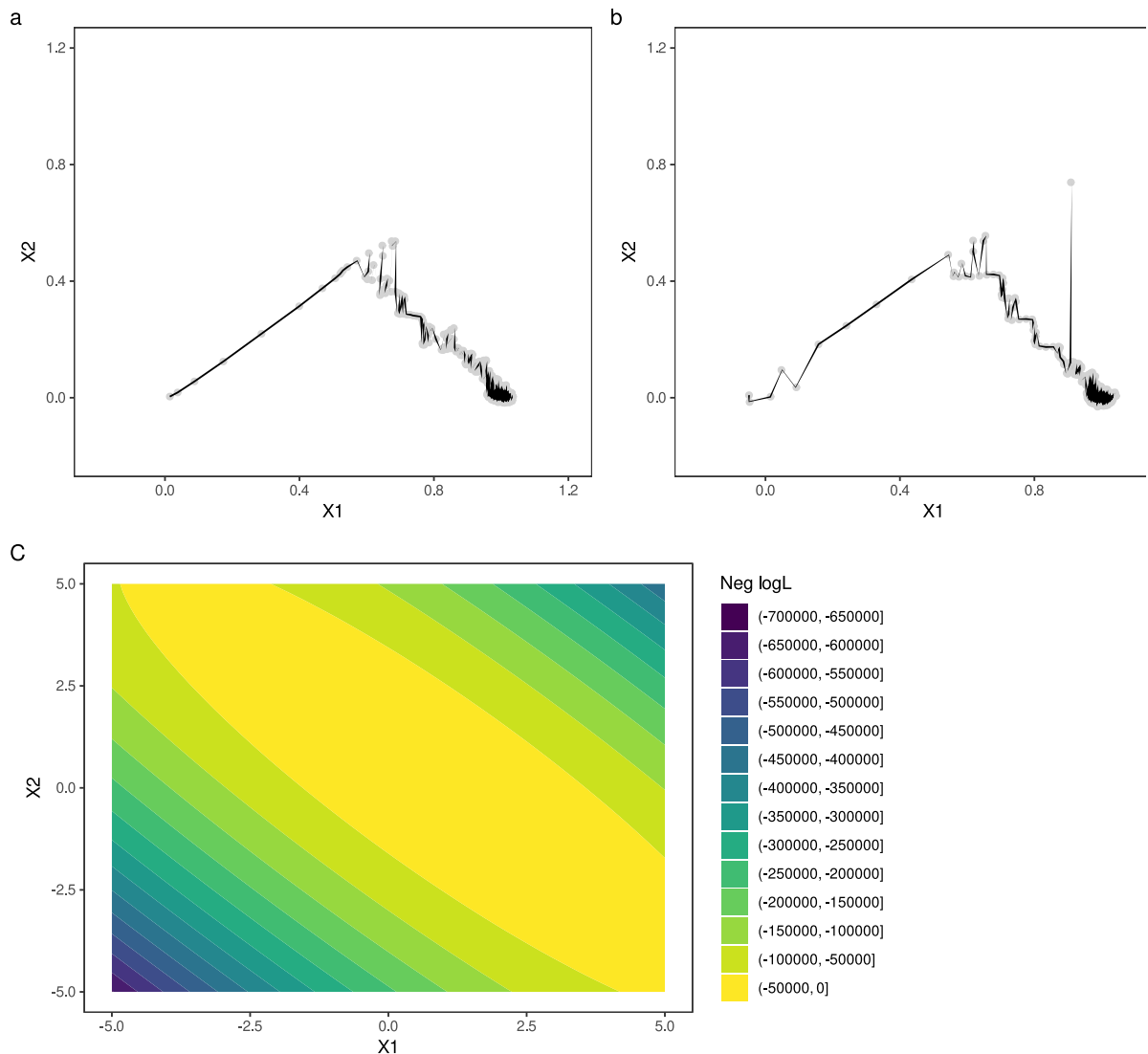

**Figure S** 13: Learning neural networks. Neural networks were trained on simulated data (1000 observations) with 5 predictors, $X_1$ has a linear effect on Y, and $X_2$ is collinear with $X_1$ (Pearson correlation factor = 0.9). The ACE was computed after each optimization step (i.e., after each batch in stochastic gradient descent) (20 repetitions). Panels A and B show the evolution of the effects for $X_1$ and $X_2$ (true effects: $X_1 = 1.0$ and $X_2 = 0.0$). Panel A shows the results for a neural network with 50 units in each of the 3 hidden layers, while Panel B shows the results for a neural network with 500 units in each of the 3 hidden layers. Panel C shows the negative log likelihood surface for the corresponding OLS.

# 8 SI References